\documentclass[review]{elsarticle}
\graphicspath{ {./figures/} }
\usepackage{float}
\usepackage{verbatim} 
\usepackage{apalike}
\restylefloat{figure}
\restylefloat{table}

\journal{Expert Systems with Applications}

\biboptions{authoryear}

\usepackage[english]{babel}
\usepackage[letterpaper,top=2cm,bottom=2cm,left=3cm,right=3cm,marginparwidth=1.75cm]{geometry}
\usepackage{amsmath}
\usepackage{amsfonts}
\usepackage{graphicx}
\usepackage[colorlinks=true, allcolors=blue]{hyperref}
\usepackage{subcaption}
\usepackage{comment}
\usepackage[T1]{fontenc}

\begin{document}

\begin{frontmatter}

\begin{center}
\vspace*{1cm}

\textbf{ \large Explainable Predictive Maintenance}

\vspace{1.5cm}

Sepideh Pashami$^{a,b}$ , S\l{}awomir Nowaczyk$^a$, 
Yuantao Fan$^a$, Jakub Jakubowski$^h$, Nuno Paiva$^{d,i}$, Narjes Davari$^d,e$, Szymon Bobek$^c$, Samaneh Jamshidi$^a$, Hamid Sarmadi$^a$, Abdallah Alabdallah$^a$, Rita P. Ribeiro$^{d,f}$, Bruno Veloso$^{d,e}$, Moamar Sayed-Mouchaweh$^g$, Lala Rajaoarisoa$^g$, Grzegorz J. Nalepa$^c$, João Gama$^{d,e}$\\
\hspace{10pt}

\begin{flushleft}
\small  
$^a$ Center for Applied Intelligent Systems Research, Halmstad University, Sweden. \{firstname.lastname\}@hh.se\\
$^b$ RISE Research Institutes of Sweden, Sweden. \{firstname.lastname\}@ri.se\\
$^c$ Jagiellonian Human-Centered Artificial Intelligence Laboratory (JAHCAI), Mark Kac Center for Complex Systems Research, and Institute of Applied Computer Science, Jagiellonian University, 31-007 Kraków, Poland. \{szymon.bobek, grzegorz.j.nalepa\}@uj.edu.pl \\
$^d$ INESC TEC, 4200-465 Porto, Portugal. \{firstname.lastname\}@inesctec.pt\\
$^e$ Faculty of Economics, University of Porto, 4200-464 Porto. Portugal. \{firstname.lastname\}@fep.up.pt\\
$^f$ Faculty of Sciences, University of Porto, 4169-007 Porto, Portugal. rpribeiro@fc.up.pt\\
$^g$ IMT Nord Europe, University of Lille, Centre for Digital Systems, F-59000, Lille, France. moamar.sayed-mouchaweh@imt-lille-douai.fr, lala.rajaoarisoa@imt-nord-europe.fr \\
$^h$ Department of Applied Computer Science, AGH Univ. of Science and Technology, 30-059 Kraków, Poland. jakub.jakubowski@arcelormittal.com\\
$^i$ NOS Comunicações S.A. 4460-191, Senhora da Hora, Portugal. nuno.paiva@nos.pt\\

\vspace{1cm}
\textbf{Corresponding Author:} \\
Sepideh Pashami \\
Center for Applied Intelligent Systems Research, Halmstad University, Sweden \\
Tel: (0046) 737552016 \\
Email: sepideh.pashami@hh.se \\

\end{flushleft}        
\end{center}

\title{Explainable Predictive Maintenance}

\author[a,b]{Sepideh Pashami \corref{cor1}}
\ead{sepideh.pashami@hh.se}

\author[a]{S\l{}awomir Nowaczyk} 
\ead{slawomir.nowaczyk@hh.se}

\author[a]{Yuantao Fan}
\ead{yuantao.fan@hh.se}

\author[h]{Jakub Jakubowski}
\ead{jakub.jakubowski@arcelormittal.com}

\author[d,i]{Nuno Paiva}
\ead{nuno.paiva@nos.pt}

\author[d,e]{Narjes Davari}
\ead{narjes.davari@inesctec.pt}

\author[c]{Szymon Bobek}, 
\ead{szymon.bobek@uj.edu.pl}

\author[a]{Samaneh Jamshidi}
\ead{samaneh.jamshidi@hh.se}

\author[a]{Hamid Sarmadi}
\ead{hamid.sarmadi@hh.se}

\author[a]{Abdallah Alabdallah}
\ead{abdallah.alabdallah@hh.se}

\author[d,f]{Rita P. Ribeiro}
\ead{rpribeiro@fc.up.pt}

\author[d,e]{Bruno Veloso}
\ead{bruno.m.veloso@inesctec.pt}

\author[g]{Moamar Sayed-Mouchaweh}
\ead{moamar.sayed-mouchaweh@imt-lille-douai.fr}

\author[g]{Lala Rajaoarisoa}
\ead{lala.rajaoarisoa@imt-nord-europe.fr}

\author[c]{Grzegorz J. Nalepa}
\ead{grzegorz.j.nalepa@uj.edu.pl}

\author[d,e]{João Gama}
\ead{jgama@fep.up.pt}




\cortext[cor1]{Corresponding author.}
\address[a]{Center for Applied Intelligent Systems Research, Halmstad University, Sweden.}
\address[b]{RISE Research Institutes of Sweden, Sweden.}
\address[c]{Jagiellonian Human-Centered Artificial Intelligence Laboratory (JAHCAI), Jagiellonian University, Poland.}
\address[d]{INESC TEC, Porto, Portugal.}
\address[e]{Faculty of Economics, University of Porto, Portugal.}
\address[f]{Faculty of Sciences, University of Porto, Portugal.}
\address[g]{IMT Nord Europe, University of Lille, Centre for Digital Systems, Lille, France.}
\address[h]{Department of Applied Computer Science, AGH Univ. of Science and Technology, Poland.}
\address[i]{NOS Comunicações S.A. 4460-191, Senhora da Hora, Portugal. nuno.paiva@nos.pt}

\begin{abstract}
Explainable Artificial Intelligence (XAI) fills the role of a critical interface fostering interactions between sophisticated intelligent systems and diverse individuals, including data scientists, domain experts, end-users, and more. It aids in deciphering the intricate internal mechanisms of ``black box'' Machine Learning (ML), rendering the reasons behind their decisions more understandable. However, current research in XAI primarily focuses on two aspects; ways to facilitate user trust, or to debug and refine the ML model. The majority of it falls short of recognising the diverse types of explanations needed in broader contexts, as different users and varied application areas necessitate solutions tailored to their specific needs.

One such domain is Predictive Maintenance (PdM), an exploding area of research under the Industry 4.0 \& 5.0 umbrella.
This position paper highlights the gap between existing XAI methodologies and the specific requirements for explanations within industrial applications, particularly the Predictive Maintenance field. Despite explainability's crucial role, this subject remains a relatively under-explored area, making this paper a pioneering attempt to bring relevant challenges to the research community's attention. We provide an overview of predictive maintenance tasks and accentuate the need and varying purposes for corresponding explanations.
We then list and describe XAI techniques commonly employed in the literature, discussing their suitability for PdM tasks.
Finally, to make the ideas and claims more concrete, we demonstrate XAI applied in four specific industrial use cases: commercial vehicles, metro trains, steel plants, and wind farms, spotlighting areas requiring further research. 



\end{abstract}

\end{frontmatter}


\section{Introduction} 

The growing complexity of Predictive Maintenance (PdM) applications in real-world scenarios, characterised by multifaceted interactions of numerous components, has brought about an increasing reliance on Artificial Intelligence (AI) solutions. The promise of these solutions is to decrease the human input necessary for building sophisticated models. Among these, black-box models developed using Machine Learning (ML), including Deep Learning (DL) techniques, have demonstrated a promising capacity for predictive precision and the modelling of complex systems.

However, the decisions made by these black-box models are often difficult for human experts to understand -- and, therefore, to act upon. The maintenance actions that must be performed based on the detected symptoms of damage or wear often require complex reasoning and planning processes involving many actors and balancing different priorities. It is unrealistic to expect the automated creation of such a comprehensive solution -- it requires too much situational context awareness. Therefore, operators, technicians and managers need insights to understand what is happening, why, and how to react.
The AI models primarily in use today, characterised by their black-box nature, fail to provide these necessary insights. They must better assist experts in grasping the ongoing situation, its causation, and appropriate responses. This is the only way to make the right maintenance decisions based on identified patterns. 
The effectiveness of a PdM system depends much less on the accuracy of the AI-triggered alerts and more on the relevance of the actions taken by operators in response.

There is yet to be a scientific or practitioner community consensus on what type of explanations are relevant for the PdM domains. Significant work must be invested into substantiating the usefulness 
of eXplainable Artificial Intelligence (XAI) in different contexts and for different actions. This paper aims to initiate a dialogue on evaluating the extent to which apt explanations of decisions made by PdM AI systems can improve outcomes across multiple facets, such as:
(i) identifying the component or part of the process where the problem has occurred; 
(ii) understanding the severity and future consequences of detected deviations; 
(iii) selecting the optimal repair and maintenance plan from a range of alternatives established based on different priorities; and
(iv) understanding the causal factors, or the reasons why the problem has occurred in the first place, for future system design improvements.
This paper pioneers a discussion in the scientific literature about the diversity of XAI explanation types beneficial for varying purposes. We provide practical examples demonstrating the variability in PdM needs across four handpicked case studies: electric vehicles, metro trains, steel plants, and wind farms.

In Section~\ref{sec:PdM-tasks}, this paper first introduces the typical PdM tasks and their associated challenges. Following this, Section~\ref{sec:purposes} delves into the reasoning behind explanations within this domain, leading to Section~\ref{sec:types}, which presents a taxonomy of current XAI techniques. Section~\ref{sec:usecases} outlines four use cases detailing specific AI system operations within the PdM domain 
 and we conclude the paper in Section~\ref{sec:conclusions}.

\subsection{Overview of XAI Research Field}

\begin{figure}[th]
    \centering
    \includegraphics[scale=0.7]{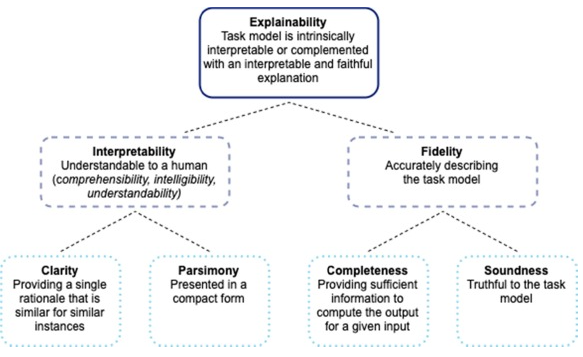}
    \caption{Taxonomy of XAI field, reproduced from \cite{2021}.}
    \label{fig:xAI-terms}
\end{figure}

Before diving into the specifics of PdM, it is necessary to provide a summary of the XAI  field to familiarise readers with fundamental concepts referenced throughout this paper. 
The foremost challenge faced by explainability in AI is to agree on an operational definition enabling us to comprehend and subsequently implement the concept. Various attempts have been made, such as \cite{MILLER20191} claiming that ``Interpretability is the degree to which a human can understand the cause of a decision,'' or \cite{NIPS2016_5680522b} stating that it ``is the degree to which a human can consistently predict the model's result,'' or by \cite{doshivelez2017rigorous} drawing from Merriam-Webster dictionary, as ``the ability to explain or to present in understandable terms to a human.''
To bring more clarity, \cite{2021} reviews the explainable AI terminology through a comprehensive survey and proposes definitions for explainability, related terms and a hierarchical structure between these, as depicted in Figure~\ref{fig:xAI-terms}.

The ability to explain is not new in AI. At least since the eighties, we can find publications \cite{mueller2019explanation} related to key concepts that identify and address challenges in understanding complex computational systems. At that time, ``Expert Systems'' such as MYCin \cite{book}, an inference engine with more than 600 handcrafted rules to support physicians in the diagnosis and treatment recommendations, were designed to interact with the users -- creating explanations through reasoning to build trust.
On the other hand, today's statistical AI methods, exemplified by Deep Neural Networks (DNNs), have revolutionised learning but often lack reasoning and explanation abilities. Long term, much research, such as Neuro-Symbolic AI \cite{garcez2020neurosymbolic}, aims to combine both benefits. Short term, it has driven a surge in XAI~\cite{Jamshidi-XAI}. 
%


Recently, \cite{VILONE202189} systematically reviewed 406 articles over the last 50 years, categorising them into reviews, notions, methods, and evaluations of AI explainability.
The requirement for AI system explanations varies significantly with the application domain, making XAI an inherently cross-disciplinary field.
Beyond surveying the current XAI landscape, \cite{MILLER20191} reviews encompassing linked areas such as philosophy, psychology, and cognitive science, investigating the human understanding and presentation of explanations.
This paper defines Human-agent interaction as the intersection of the artificial intelligence, social sciences and human-computer interaction.
Acknowledging this multidisciplinarity, the field of Human-Computer Interaction also delves into how Data Scientists understand Machine Learning models through interactive interfaces~\cite{10.1145/3290605.3300809}, or how they over-trust and misuse existing interpretability tools~\cite{10.1145/3313831.3376219}. 
For instance, \cite{liao2021questiondriven} recommends a question-driven design process to enhance stakeholder collaboration on AI/ML products, tackling diverse explainability goals such as system improvement, compliance control, decision justification, and discovery of transferable new patterns \cite{belle2020principles}. 
The notion of requiring different explanations for different stakeholders is also endorsed by \cite{https://doi.org/10.48550/arxiv.1802.01933}, based not only on competencies by also user's available time. 




\section{Predictive Maintenance Tasks}
\label{sec:PdM-tasks}

This section breaks down key concepts tied to the broadly defined field of Predictive Maintenance (PdM). Notably, we adopt a relatively comprehensive definition of PdM, given the interconnectedness of these tasks. AI systems typically progress from more straightforward to complex configurations, with explainability often pivotal in this evolution, as the activities that used to be done manually are becoming automated to a more considerable and significant degree.

Currently, various industries employ black-box AI systems to monitor operations and predict failures based on analysing sensor data. These systems identify signs of impending issues by detecting anomalies and departures from normal behaviour, often with impressive accuracy. 
However, PdM operates within a broader context, striving to pinpoint likely causes and formulate optimal solutions before a problem escalates. In complex systems, merely knowing something is wrong (i.e., detecting an anomaly) is insufficient; the key lies in discerning its underlying reasons and potential consequences in order to offer solutions or advice to counteract those consequences. However, that necessitates intricate interplay among various industrial operations~\cite{Nowaczyk2020}.
The entire decision chain can seldom be fully automated, despite recent advances in AI; in practice, multiple small and isolated AI/ML models are typically deployed, often used by different teams of humans. 
For example, repairing a wind power plant requires coordinating inventory (availability of replacement parts), logistics (securing transport to the site), personnel management (staff availability), and weather forecasts (assessing suitable conditions for maintenance, especially for tall towers in offshore farms), among other factors. AI alone cannot manage these tasks; they demand human expertise since the full relevant context cannot be sufficiently formalised to permit automated reasoning~\cite{Bae2020Interactive}.



\begin{figure}[ht]
    \includegraphics[width=0.99\columnwidth]{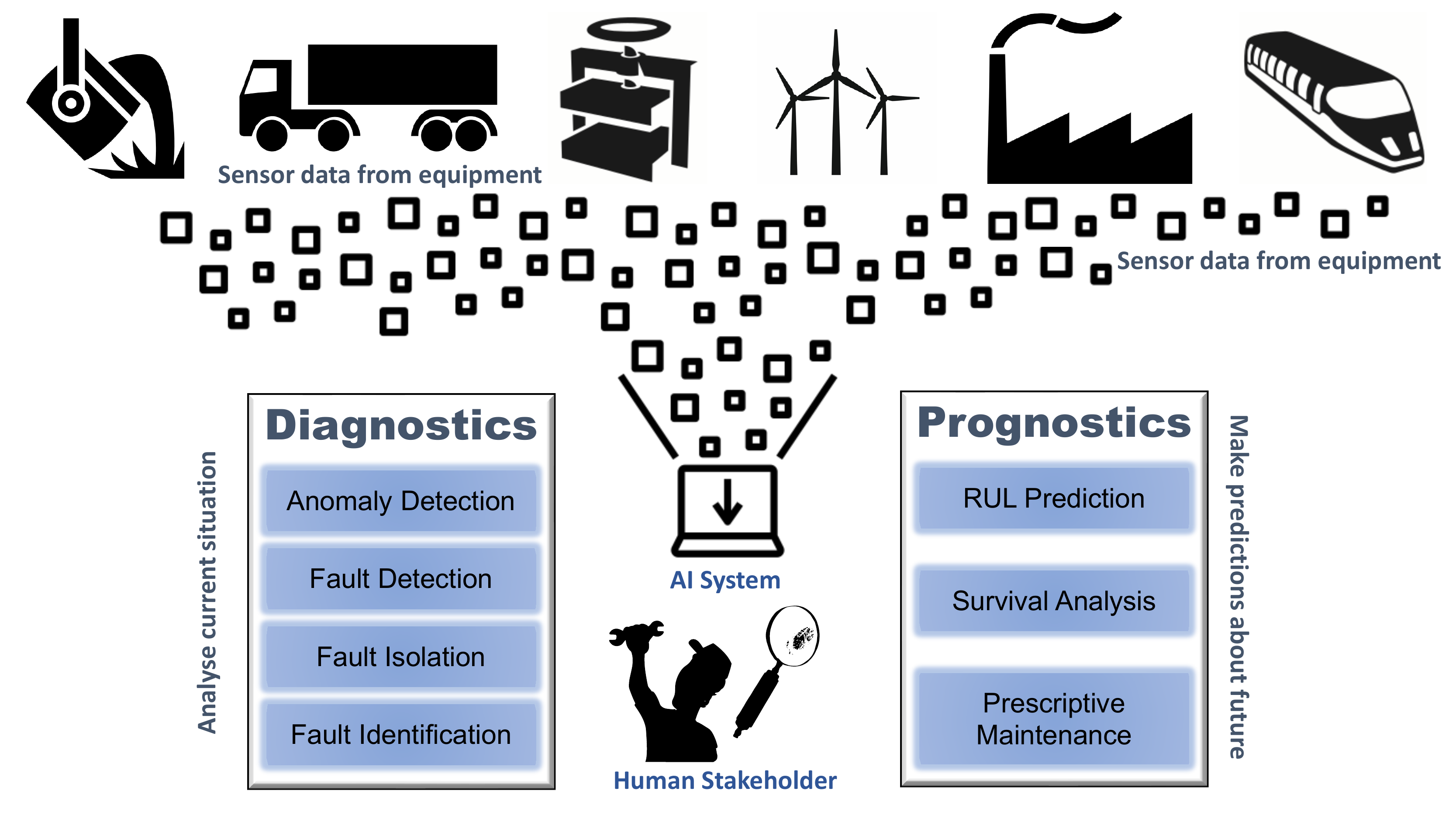}
    \caption{Overview of key predictive maintenance tasks.}
    \label{fig:PdM_Tasks}
\end{figure}

Conceptually, Predictive Maintenance strives to proactively mitigate potential or identified faults based on the condition of the equipment, thereby ensuring its optimal operation. Figure \ref{fig:PdM_Tasks} depicts essential PdM tasks. While it is tempting to view PdM as merely relating to the ``future'' (right-hand side of the figure), such a view is somewhat restrictive, as prognostics cannot function independently of diagnostics. Especially concerning explainability, these tasks must be tackled collectively. The later solutions cannot be interpretable if the initial stages are executed incomprehensibly.

Using collected data, PdM monitors equipment to maintain a ``healthy'' state -- one desired for the operation. Any significant deviation from this norm is considered an ``anomaly'' (or an outlier). A ``fault'' is any unpermitted deviation, i.e., once the equipment moves away from what is \textit{acceptable}.
If a fault permanently disrupts the system's function, it becomes a ``failure.'' 
Historically, the most natural maintenance strategy is the ``reactive'' one, i.e., repairing equipment after failure. 
A relatively obvious alternative is ``preventive,'' which uses periodic repairs or replacements (based on population-level statistics), balancing avoiding failures versus wasting resources. 
The goal of ``predictive maintenance'' is to
accurately determine the exact condition of each in-service equipment and perform maintenance at the exact right time. That perfect moment is identified by analysing component or subsystem ``degradation curves,'' i.e., trajectories of  health indicators for a particular failure mode.

Model-driven methods rely on the physical properties or system models, offering high reliability and safety but requiring extensive skilled labour. Examples include capturing the dynamics or kinematics of moving parts, the effects cracks of various sized and shapes have on measured vibrations, and phenomena of partial discharge that happen in electrical circuits.
Data-driven methods use historical data to autonomously identify relationships between sensor data and outcomes, offering a low-cost development but often without performance guarantees. They are extensively applied in highly complex and uncertain areas, for example, chemical and metallurgy processes, vehicles and transportation systems, or intelligent buildings.
In practice, AI systems usually blend expert knowledge and collected data, creating hybrid approaches. This paper focuses on works with significant data-driven elements, either pure data-driven solutions or relevant hybrids.

In this context, all PdM tasks are based on exploiting past data. The primary division in the maintenance decision-making process is between diagnostics and prognostics, based on the time horizon of interest: diagnostics is interested in understanding what is happening \textit{right now}, while prognostics looks into the \textit{future}.
In the rest of this section, we elaborate on how these crucial tasks exploit data to yield valuable insights into a system's operation and help take the correct action. We define each task, and the motivation for its usefulness in the PdM context, list typical challenges encountered in practice, give examples of state-of-the-art methods, and end with open questions from the explainability perspective.

\subsection{Diagnostics}

Diagnostics involves identifying the equipment's current state or condition and assessing any pre-existing damage. Real-time data and continuous monitoring facilitate the right actions to be taken at the right time.
Diagnostics consists of anomaly detection, fault detection, fault isolation, and fault identification (also called fault diagnosis). Each task builds upon the preceding one, yielding more nuanced information. Initially, anomaly detection pinpoints any operation deviating from the norm. Subsequently, fault detection establishes whether it is a fault, as not every deviation indicates a problem. The next stage, fault isolation, seeks to locate the fault, e.g., determining which component is responsible. Finally, fault identification estimates the fault's severity, such as the magnitude of a crack or leak.

On a high level, the usefulness of XAI for diagnostics chiefly concerns supporting human actors in the follow-up or downstream tasks -- enabling decision-making that extends beyond the scope of the AI system. 
Naturally, this is in conjunction with the universal advantages XAI offers across virtually all AI tasks, such as fostering trust in the AI system, aiding in the debugging and refinement of the model, and fulfilling legal requirements.


\subsubsection{Anomaly Detection}


Anomaly detection is a process that seeks out patterns or events in data that deviate significantly from what is considered normal or expected. It assigns anomaly scores to any given data point or time period, measuring the degree of abnormality. This measure is based on comparing these data points against the most typical historical data of the equipment in question~\cite{aminikhanghahi2017survey} or against a fleet of similar equipment~\cite{rognvaldsson2018self}. Anomaly detection is mainly studied in an unsupervised ML setting, without explicit labels -- instead based on the assumption that anomalies are rare events, and the majority of data is normal.


Anomaly detection plays a crucial role in modern industrial processes, which rely heavily on extensive sensor networks for continuously monitoring facilities, systems, and equipment. Given the sheer volume of data collected, manual analysis is unfeasible. It is also often impossible to anticipate all the possible faults. As such, anomaly detection algorithms provide a vital tool to alert operators to unusual behaviour in the machinery. While such situations may be associated with atypical usage or unique external conditions, they offer early warnings of potential issues. This can prompt maintenance staff to take preventative actions, thereby minimising downtime and improving both the reliability and efficiency of industrial processes.

Industrial data analysis, however, presents complex challenges. These include capturing temporal dependencies between observations, including long-term dependencies and future consequences of past actions; addressing the nonstationary nature of industrial data, as the statistical properties change over time, e.g., based on seasonality or economic trends; handling data quality issues due to noise, missing information, outliers, and explicit bias; and dealing with high-dimensional and imbalanced time-series data exhibiting complex interactions between parameters and scarce examples of anomalies.
Another significant challenge is the difficulty in evaluating models due to their unsupervised nature, inherent subjectivity, and domain dependency on what constitutes a true or false positive.

Many techniques have been proposed for anomaly detection, and XAI promises to address at least some of these challenges in industrial processes, as seen in~\cite{chatzigiannakis2006hierarchical, serradilla2021adaptable}. 
In recent years, deep learning~\cite{pang2021deep,li2022deep} approaches have shown promising results. They are frequently constructed as black boxes whose outputs as scalar anomaly score that interprets the degree of abnormality for specific data instances, including unsupervised ones \cite{munir2018deepant, audibert2020usad}. However, these deep anomaly detection models are often criticised for their lack of explainability; relatively little research is exploring this issue. As two examples, \cite{siddiqui2019sequential} is based on a search in a subset of features that makes a reported anomaly most abnormal, while \cite{myrtakis2021comparative} focuses on explaining the outlying-ness features of multi-dimensional data points to give the best explanation on why an outlier deviates from the inliers. In~\cite{CALIKUS2020113453}, authors propose a SAFARI framework created by abstracting and unifying the fundamental tasks within the streaming anomaly detection.


While anomaly detection has been a subject of extensive research, there is a scarcity of studies that focus on explanation techniques for anomalies, with only a handful of surveys~\cite{li2022survey, sejr2021explainable, pang2021deep}. Explainable Predictive Maintenance is a highly relevant case study for anomaly detection. Offering insights into why an anomaly has been triggered and aiding experts in understanding and addressing the underlying causes and effects that remain largely unsolved. The extent to which these techniques can be integrated and scaled in real-world applications remains an open research area.

\subsubsection{Fault Detection}


Fault detection is a process that identifies whether faults are present within a system. A fault is an anomaly or deviation that goes beyond the desired operating condition of the equipment as per its specifications. Data anomalies in signal readings often indicate the possibility of one or more subsystems operating in a faulty state. 
Distinguishing an operation that is simply unusual from one that indicates early symptoms of a problem is crucially important, especially in complex systems comprising multiple components. 

The principal reason for implementing fault detection methods in industrial processes is to warn about defective system operation changes promptly \cite{rajaifac, alves}. Early fault detection, while the system is still running in an acceptable state, allows maintenance to prevent problem escalation. 
If unchecked and uncorrected, faults usually evolve into failures, leading to a complete halt in equipment operation. 

In the industrial setting, however, fault detection presents several challenges. These include the ability to detect issues as quickly as possible to enable acting promptly; distinguishing between various types of faults and separating them from simply atypical operations; demonstrating resilience to different kinds of noises and the uncertainties inherent in the input data; adapting to both short-term and long-term changes in the environment and external conditions. One of the most significant challenges is finding the right balance in the timing of alerts; these should be early enough to allow time for intervention and prevent failures, yet not so premature as to cause false alarms and incur unnecessary costs.

Due to its adaptability and precision, data-driven fault detection has become increasingly popular for monitoring complex industrial processes. The growth and effectiveness of this approach are demonstrated in several surveys presenting existing techniques; see~\cite {venkatasubramanian2003review, lei2020applications}. 
While most approaches use traditional black-box fault detection methods, the advantages of XAI are discussed in several recent works~\cite{mey2022explainable,madhikermi2019explainable,srinivasan2021explainable} for promoting user understanding and enhancing trust. Furthermore, detailed explanations about faults facilitate more efficient execution of subsequent tasks such as fault isolation and identification \cite{bhakte2022explainable,brito2022explainable,grezmak2019explainable,serradilla2021adaptable}.

While XAI for PdM promises to address many limitations of black-box fault detection models, questions remain regarding the most effective methods to establish user trust and deliver comprehensive, easily understandable explanations regarding fault detection and its associated symptoms.
Aiding users in comprehending the rationale behind detected faults is expected to offer valuable insights into fault symptoms, enabling the effective execution of subsequent tasks.

\subsubsection{Fault isolation}



Fault isolation is the process conducted after a fault has been detected based on observable symptoms or unusual behaviour. The purpose of this process is to identify the nature and location of the fault. Defects can manifest due to various causes, including degradation over time, accidents, misuse, or functioning under unfavourable external conditions. Different faults may have similar symptoms, making it non-trivial to isolate the actual underlying cause of any encountered problem.

While faults can sometimes be tolerated in their early stages, diagnosing and addressing them proactively through maintenance actions is crucial. In order to avoid potentially very severe consequences of a fault, the component or subsystem responsible needs to be repaired or replaced. This can only be done if the underlying cause of the fault is accurately identified. Mistakes at this stage often manifest as replacing healthy components, not only incurring additional maintenance costs but also -- ultimately -- failing to prevent the imminent failure.

%

The industrial setting, however, comes with various challenges. The effectiveness of fault isolation methods often hinges on an extensive understanding and expertise within the specific application domain \cite{migueis2022automatic}. Building this knowledge can be resource-intensive and time-consuming. Moreover, these methods may not be readily generalisable to new problems and may not be cost-effective when dealing with complex systems.
Fault isolation needs to be able to capture the key discriminating factors between different faults, even if the resulting symptoms are similar.


Various data-driven approaches exist for fault isolation, including multi-class classification \cite{dineva2019fault,pang2020spatio}, clustering \cite{may2021multi,serradilla2021adaptable}, and multiple hypotheses testing \cite{basseville2002fault}. However, most of these methods are ``black-box'' and do not provide users with explanations about their predictions. Root Cause Analysis (RCA) and system identification-based methods are other common approaches to fault isolation \cite{hwang2009survey,serdio2015fuzzy,forrai2016system}. Recent research has focused on incorporating explainability in fault isolation~\cite{grezmak2019explainable,costa2022fault}, often using patterns in frequency spectrums to indicate different fault modes.

From the user's perspective, it is crucially important to understand the differentiating factors between different faults. The XAI solutions promise to provide more detailed information regarding the fault, making it easier to act upon. The connection between the corresponding component or subsystem and the observed symptoms is also meaningful to launching further investigation evaluating the severity level of the fault, and scheduling the maintenance accordingly. Understanding the root cause of a fault, and potential relationships between fault modes, can be beneficial for improving equipment design.

%
%
%

\subsubsection{Fault identification}

Fault identification aims to determine the size or severity of the fault, as well as its time-variant behaviour. It quantifies the extent of a fault's impact on a system, thus permitting an accurate measure of its potential to interfere with regular operations. Sensor data is analysed to understand the underlying characteristics of the fault, such as its magnitude or frequency of occurrence. It also considers the time-variant behaviour of a fault, which refers to changes in those characteristics over time. Fault identification provides detailed insight into a system's current operational status, ultimately facilitating the maintenance team's decision-making process.

The significance of fault identification in PdM revolves around its capability to assess the severity of detected faults accurately. By understanding the fault severity, one can design the right maintenance measures, including decisions such as whether to repair or replace the affected component, or how urgently an action is needed. Thus, maintenance can be scheduled and implemented effectively, avoiding unnecessary shutdowns but also ineffective or overly aggressive repairs. 



Fault identification in complex industrial applications is challenging mainly due to issues with sparse and imbalanced data, along with unreliable or faulty sensors. 
These issues often result in poor generalisation performance and low accuracy of the model predictions. Since faults are rare events, the AI system has seldom been trained on a representative sample of them. Furthermore, fault identification techniques tend to struggle with non-stationarity in the systems. They heavily rely on the available fault data, which is often collected under idealised conditions and not representative of real-world scenarios. Such limitations underscore the need for hybrid methodologies that blend model-based and data-driven strategies. For instance, \cite{piltan2021crack} provide a robust method for fault diagnosis and crack size identification in bearings combining the advantages of both techniques.

There are several noteworthy approaches to fault identification. One class of methods uses a residual signal to locate and quantify the magnitude and severity of a defect. This approach is beneficial when the fault's impact on the system is not entirely known; \cite{zhang2022intelligent} reviewed a number of relevant papers, dividing them into three categories: data augmentation-based, feature learning-based, and classifier design-based. 
Other researchers~\cite{lei2009gear,chen2013fault, liu2018artificial,qiu2019deep} have used supervised ML regression or classification algorithms. Meanwhile, \cite{lundgren2020data} employed optimisation techniques for estimating fault severity, demonstrating their potential for effective fault identification.

As predictive maintenance evolves, new questions arise in terms of explainability. The ability to extract meaningful patterns from raw data is crucial for effective fault identification. These patterns can provide essential insights into the fundamental dynamics of the system and the underlying causes of faults. The application of semi-supervised models presents an interesting opportunity, potentially allowing for the classification of faulty data instances based on their similarities, thereby identifying different types of faults. These insights could then be used to shed light on the plausible reasons for each fault type. Despite the advancements in this field, more research is required to maximise the benefits of fault identification in predictive maintenance, particularly from an explainability perspective.

\subsection{Prognostics}

Unlike diagnostics, which concerns analysing the current health state of the equipment, prognostics is concerned with the future condition, i.e., the damage that has not yet occurred. It focuses on forecasting the system's degradation over its future lifetime and evaluating how the risk of failure changes over time. In other words, prognostics determines whether a failure is impending and estimates how soon it will occur.

Typical tasks for prognostics include Remaining Useful Life (RUL) estimation, survival analysis and prescriptive maintenance. RUL estimation predicts the time left until the equipment no longer functions according to the specification; a particularly relevant special case is failure prediction, namely forecasting whether a failure will occur within a pre-determined time frame. Survival analysis, also called reliability theory, studies -- on a population level -- the probability distribution of times until an event of interest, such as a failure, occurs.
Finally, prescriptive maintenance expands beyond individual predictions and concerns a complete asset maintenance strategy, adjusting operating conditions for desired outcomes through scheduling and planning.

In broad terms, the value of XAI in the context of prognostics primarily lies in its ability to guide human stakeholders in the planning and execution of future operations. This includes accounting for diverse scenarios, where explanations help decision-makers understand where the AI system has already considered a particular factor, or if they need to make adjustments accordingly. It is also vital to understand the limitations of the current, potentially deteriorated, condition and adjust the short-term equipment usage accordingly, considering how the wear progression may vary under different circumstances. Finally, explainable prognostics allows for understanding the key factors that influence failures, degradation, and wear, which is vital for the enhancement of future designs. Naturally, as with diagnostics, aspects like building trust, improving models, and satisfying legal requirements retain their importance.

\subsubsection{Remaining Useful Life Estimation}

The Remaining Useful Life (RUL) of a system is defined as the time interval from the particular time of operation until the end of the system's useful life, i.e., when it is incapable of performing its functions \cite{si2011remaining}. Predicting RUL is commonly considered as learning a functional mapping between observations (sensor measurements) and the health condition of the equipment, followed by predicting the future deterioration of that health.
In other words, RUL estimation aims to predict the future functional state of a system; it provides the length of time a machine or component is expected to continue performing its intended function before failure. This prediction is typically based on the current state of the system and its degradation trends.

RUL estimation is an essential part of predictive maintenance strategies, as it helps decide the best time to conduct maintenance, thus avoiding unnecessary downtime or catastrophic system failures. Routine preventive maintenance leads to excessive and unnecessary costs when parts are replaced prematurely. On the other hand, waiting for a component to fail before replacing it can lead to even more significant costs associated with downtime, repair, and potential secondary damage. RUL estimation allows for a more effective maintenance strategy, where components are replaced or repaired just before they are likely to fail. This not only reduces costs but also optimises resource allocation. RUL estimates also contribute to strategic decisions about asset management, such as when to retire or replace equipment, how to optimise the use of current assets, or when to invest in new technology.

Estimating RUL is often challenging due to uncertainties and complexities in degradation processes, operational conditions, and variations in material properties. RUL estimation depends on the availability of high-quality historical data, which includes records of system operation, maintenance, and failures. However, such data may be scarce, incomplete, noisy, or inconsistent in many real-world scenarios. This makes it difficult to train reliable models for RUL prediction.
In addition, degradation processes are influenced by many factors, such as usage patterns, environmental conditions, and inherent variability in material properties. This makes the degradation process uncertain and stochastic in nature, leading to considerable variability in the RUL of identical systems under similar operating conditions. It can be challenging to model this variability and uncertainty effectively.
Finally, modern systems and machinery often have complex structures with intricate interactions among different components, adding an additional layer of complexity. 
Validating RUL predictions is inherently difficult because it requires waiting until the end of life of the system to compare the actual and predicted RUL. This can take a long time, particularly for systems designed to last for many years.

In general, there are three types of approaches for RUL prediction: i) mapping between a set of sensor inputs and RUL (e.g. neural network-based methods \cite{rigamonti2016echo,zheng2017long,remadna2020leveraging,Chen2019}); ii) mapping between an approximated health index or degradation level and RUL (e.g. \cite{liu2013data,le2013remaining,zhao2017remaining,fan2020transfer}); iii) Similarity-based matching with or without training a mapping function (e.g. \cite{wang2008similarity,wang2010trajectory}).
Conventionally, RUL prediction methods in the current industrial assets were often developed based on controlled experiments with simulated operating conditions (e.g., stress test) and pre-defined faults \cite{li2014experiments, nectoux2012pronostia,kleyner2017new}.
It assumes data from the simulation at the training time, and the unseen future data are of the same population. However, this assumption might not hold for complex machines that can do different things and operate under varying conditions. They might deteriorate in unexpected ways, especially if new operational settings and conditions appear post-deployment of the equipment.
The chosen approach depends on various factors, such as the availability of data, the complexity of the system, and the desired level of accuracy. Regardless of the approach, the goal remains the same: to predict the RUL accurately, allowing for optimal planning of maintenance activities, improved operational efficiency, and reduced costs.

A particularly interesting special case of RUL estimation is "Failure Prediction", an alternative approach to forecast whether any component or system failure would occur in a pre-defined time, referred to as the prediction horizon \cite{PRYTZ2015139}. Ideally, the failure should be detected early enough so maintenance can be scheduled to fix the problem. Work \cite{fan2015evaluation} on forecasting air system failures in city buses proposed a three-month prediction horizon. Failure prediction can be formulated as a classification task, where the observations within the prediction horizon (before failure) were considered faulty, and failure-free observations were considered healthy. For dealing with multiple types of failure, faulty samples are assigned to a specific fault category, and healthy samples are ``shared'' between all fault categories.

%
Black-box methods, such as Deep Learning algorithms, have the potential to capture intricate relationships in data for RUL estimation accurately. However, they often lack transparency, making comprehending the reasoning behind their predictions challenging. This opacity can trigger trust issues, especially in areas where safety is paramount, and stakeholders require insight into the prediction process.
To address this, at an individual prediction level, XAI can provide localised explanations that contextualise the model's prediction. This is particularly valuable for decision-makers who must consider a multitude of factors when choosing the appropriate maintenance action. By providing clear explanations, XAI can aid operators, technicians, and managers in understanding how these factors interact with the model's output.
On a broader scale, XAI can provide global explanations, offering valuable insights into significant factors that influence the deterioration process, such as varying operating conditions or usage patterns.



\subsubsection{Survival analysis}
Survival analysis calculates the probability distribution of survival of humans, equipment, machines, or systems. In the predictive maintenance context, survival models calculate the probability of time to a breakdown of the equipment and estimate how equipment degrades during its lifetime. These models can also treat the early replacement of equipment and replacement due to breakdown differently. 

In the context of PdM, survival models are particularly interesting since they consider a ``population-level'' perspective. Namely, instead of looking at an individual system in operation, and focusing on its unique features, they consider that unit a part of a bigger group. 
The specific time when it fails, then, becomes just one instantiation of a more complex random variable. This ability to consider the inherent randomness of the equipment failures allows abstracting away from specifics and focusing on more high-level concepts.
Understanding the strong and weak points of any given design necessarily takes this kind of population-level perspective.


One substantial benefit of survival models is that they can take advantage of censored data in their analyses which means they consider the partially observed data of the equipment’s operations even if data is unavailable during the whole study. Additionally, survival models can often predict the risk factors for individual equipment or groups of equipment in a way that can identify a high or low-risk concerning failure occurrence. 
Traditionally, statistical methods such as the Cox Proportional Hazard model \cite{cox} and Kaplan-Meier estimator \cite{Kaplan_Meier} have been widely used for survival analysis. Although the successes of these methods come from their simplicity and ease of implementation, the drawbacks often come from their strong assumptions and challenges in handling non-linearity. Nowadays, due to data availability, variations of machine learning models have been developed to model survival functions. Random Survival Forests (RSF) \cite{RSF} is an extension of Random Forest, which is an ensemble of multiple survival trees. Deep generative models can be utilised to model time to event \cite{date, pmlr-v56-Ranganath16}.

In the survival analysis context, it is desirable for XPdM to help, for example, in understanding (i) what are the factors that lead to better or worse survival time; (ii) what are the discriminative factors between low and high risk of failure for different groups and, (iii) what are the minimum actions which can be taken to remove a piece of equipment from a high-risk group?  

\subsubsection{Prescriptive maintenance}
Prescriptive maintenance has been recently introduced as the next evolution of maintenance strategies, to follow predictive maintenance concepts~\cite{nemeth_prima-x_2018,meissner_developing_2021}. 
In this type of strategy, the AI algorithms are not only making the predictions, but they also control the system directly and perform planning for addressing faults and failures~\cite{matyas_procedural_2017,ansari_prima_2019}.
An effective prescriptive maintenance approach is expected to be superior to the greedy usage of resources that arises from reacting to predictive maintenance alarms without paying attention to operational costs and constraints. For example, when managing a fleet of aeroplanes, keeping the schedules and avoiding delays is highly important \cite{meissner_concept_2021}. In this scenario, 
making repairs based solely on suggestions from even the best predictive models might lead to long flight delays. Another example would be a chemical complex, where safety and prevention of accidents are crucial \cite{gordon_data-driven_2022}. A greedy approach to maintenance might lead to an overcommitment of resources and, thus, a higher risk of relatively minor incidents turning into dangerous situations. Key factors that should be incorporated, depending on various applications, are personnel scheduling, efficient ordering of replaceable units, and even costs related to labour, material, and emissions \cite{choubey_holistic_2021,meissner_developing_2021}.

Prescriptive maintenance is typically implemented as an algorithm that optimises the total cost consisting of multiple factors. The output of the algorithm is a plan, or a course of action, including resource management and decision-support or recommendation systems \cite{ansari_prima_2019}. The algorithm continuously searches for the best strategy while taking input from the predictive models and other online sources of information. In principle, the output strategy could be determined entirely automatically, as opposed to decision support systems that necessitate a human in the loop.

One of the main challenges with this type of strategy is formulating the optimisation problem. Determining the factors that should be included in optimisation and how they should be balanced is domain-specific and difficult. Thus, incorporating expert knowledge is essential, and purely data-driven designs remain infeasible. 

Since, in principle, prescriptive maintenance allows for complete automation, there is not much work relating it to XAI. We were unable to find any prescriptive maintenance work in the literature that takes advantage of explainable predictive solutions. 
However, explainable predictive models are essential for establishing trust in the prescriptive maintenance solution.
They could also give more insight into the interdependencies in the machinery to reduce the overall long-term cost by focusing on replacing more critical parts.

\section{Purposes of Explanations in Predictive Maintenance}
\label{sec:purposes}

Explainable predictive maintenance can, on a high level, enhance the comprehension of a product or process's design and usage. It offers additional insights and corroborates domain experts' existing knowledge and intuition. This understanding aids in refining the development of data- and knowledge-driven methods, presenting data scientists with more tools to troubleshoot and assess their models beyond standard performance indicators like accuracy. Lastly, explainability can aid decision-making processes. It raises awareness about the gravity of the situation, points out necessary supplementary actions, and builds confidence in the PdM system. This way, it ensures that the decisions made are more informed, precise, and reliable, enhancing the overall efficacy of the PdM system.

Integrating AI systems into industrial practice presents distinct challenges, specifically in crafting maintenance plans. These primarily revolve around incorporating AI-generated outputs into human decision-making processes and harmonising them with existing human expertise. It is vital that the AI outputs, such as fault predictions, are contextualised in a manner that is relevant and understandable to human users, a prerequisite to making AI valuable and reliable.

To this end, providing explanations tailored to different stakeholders' roles and needs is crucial. For instance, maintenance engineers may need explanations that directly relate to the technical blueprints of the installation they handle, enabling them to understand and implement the AI's recommendations accurately. On the other hand, managers tasked with assessing the financial implications of downtime require a business-focused explanation that presents potential costs and savings clearly and concisely.

Moreover, other stakeholders, such as company lawyers, might need a different approach altogether. They might require information that helps them assess potential liability issues in the event of a safety-threatening failure. Hence, the interactions between human experts and AI systems in industrial PdM settings are multifaceted and need careful orchestration to ensure the effective and meaningful integration of AI outputs into human decision-making processes.

In stark contrast, in today's XAI practice, the underlying theme of the XAI is often exclusively either building up trust towards AI systems, or supporting data scientists in model debugging and refinement. While these purposes indeed remain relevant, as we have indicated earlier in this paper, they are far from exhaustive -- and, most likely, not even the most important ones.

In order to maximise PdM's actual usefulness, we need to go one step further. In particular, the research community needs to develop technological solutions that explicitly address the four broad reasons why explanations are required for PdM. Within these four rationales, the explanations of AI decisions will lead to the most significant improvement in human experts' repair and maintenance actions. There is a common theme to all of them: while, in theory, the scope of the AI PdM system could be expanded to cover those needs and make interactions with humans unnecessary, in practice, today's isolated and specialised solutions cannot achieve such autonomy. Until we develop Artificial General Intelligence, these key interfaces between humans and machines remain challenging.

The first aforementioned rationale is identifying (isolating and characterising) the fault. In complex industrial systems, AI can often detect deviations from normal behaviour. However, due to many components with complex interactions, pinpointing the exact culprit requires engaging experts and their domain knowledge. The second rationale is to understand fault consequences. Proper maintenance depends on how the issue will evolve, what is the Remaining Useful Life, but also what can be the collateral damage or associated loss of productivity. These questions, again, require considering the broader context, beyond the scope of most deployed AI systems. The third rationale is supporting domain experts by helping human operators create the proper maintenance or repair plan. It includes optimising the system performances (e.g., uptime or safety) in the presence of the degradation until the problem can be removed, to strike a suitable trade-off between different quality criteria. Finally, the fourth and final rationale is about understanding why the fault occurred in the first place and how to improve the system in the future. It can concern incorrect usage, suboptimal design, and the monitoring process itself. For example, what optimisation of sensor types and placement would allow for earlier detection in the future, or what changes to the manufacturing process parameters will extend the lifetime of specific critical components?

It can be seen that all four of these concerns share a core difficulty: they cannot be solved by the expert nor by the predictive maintenance AI system alone. The proper maintenance solution can be created only in collaboration between the two. Therefore, there is a need to develop novel context-aware XPdM decision support modules, using the explanations from the AI system, to make this collaborative process between all the actors feasible and efficient.


As a somewhat unique feature, it is worth noting that predictive maintenance explanations are, naturally, rarely needed for normal cases. Instead, explanations are primarily used to rationalise why machine learning predicts an imminent failure or considers current operations as abnormal. In particular, the connection between the predictions and the underlying physics governing the behaviour of the system can be leveraged for an explanation, which is generally assumed to be well-understood by the expert and can therefore provide common ground between the AI and the human.  


There is a heated discussion in the literature today \cite{Rudin2019} about the correct outlook towards XAI: should it be based on an extra layer to be added on top of existing black-box models or achieved through the development of inherently interpretable (glass) models. Much work is being done in both these areas \cite{molnar2020interpretable}. In the PdM domain, both approaches have merit, but there are too few honest comparisons between them. There is also a need for more specific evaluation metrics, especially Functionality- and Human-Grounded ones \cite{doshi2017towards}. In particular, ones that capture the needs of different actors, based on their competence levels and specific goals, since within any given industry, multiple stakeholders must interact with a PdM system, often for various reasons. Understanding their demands and making sure each receives the proper support is critical.


\subsection{Model Development and Evaluation}  
The typical machine learning pipeline, which also applies to predictive maintenance tasks, includes the following steps: data filtering, feature engineering, model definition, hyperparameter tuning and model evaluation.
Most predictive maintenance models heavily rely on training data; therefore, the quality of the data used in the development of the model plays a crucial role in the final accuracy of the solution. 
In PdM tasks, sensor data is usually used, which is much harder to evaluate manually than image or text data, especially in unsupervised problems.
This issue should be addressed at the pre-modelling stage, which was described in more detail in Section \ref{sec:premodelling}.
Another critical element is the proper selection and optimisation of the machine learning algorithm.
Most machine learning algorithms, especially complex ones (e.g. deep learning, random forests, support vector machines), are not interpretable, meaning the developer has no inherent understanding of what the model actually learned.
Any issues that are not discovered and resolved early in the ML pipeline will usually negatively impact model performance.
Without an understanding of the model, the developer may not be able to spot weak spots in the developed pipeline and therefore end up with a solution far from optimal.
This is one of the reasons why model understanding is crucial to improve the reliability and accuracy of the developed solutions. 

To account for this issue, explainable AI methods may be used to discover the trends within the PdM model so that the developer possesses more information on the trained model behaviour.
For example, if we collected yearly data from a manufacturing process when specific faults occurred only in winter, then a model could be easily biased towards the ambient temperature, and thus could not correctly predict the same failure in summer.
A different scenario in which XAI might improve model development is to check its validity against background knowledge.
In the case of correlated features, the model can often learn based on one of the dependent variables instead of an independent one, which may lower the model's performance in production.
Another potential issue, which may be discovered using XAI techniques, is the situation when the black-box model points out different features (usually correlated) for predicting similar observations.
For example, the PdM model predicts the RUL of a component, and to predict RUL=30 days, it relies on Feature 1, while to predict RUL = 29 days, it relies on Feature 2. Thus, the model might be unreliable due to its low stability.

\begin{figure}[ht]
    \centering
    \includegraphics[width=1.0\columnwidth]{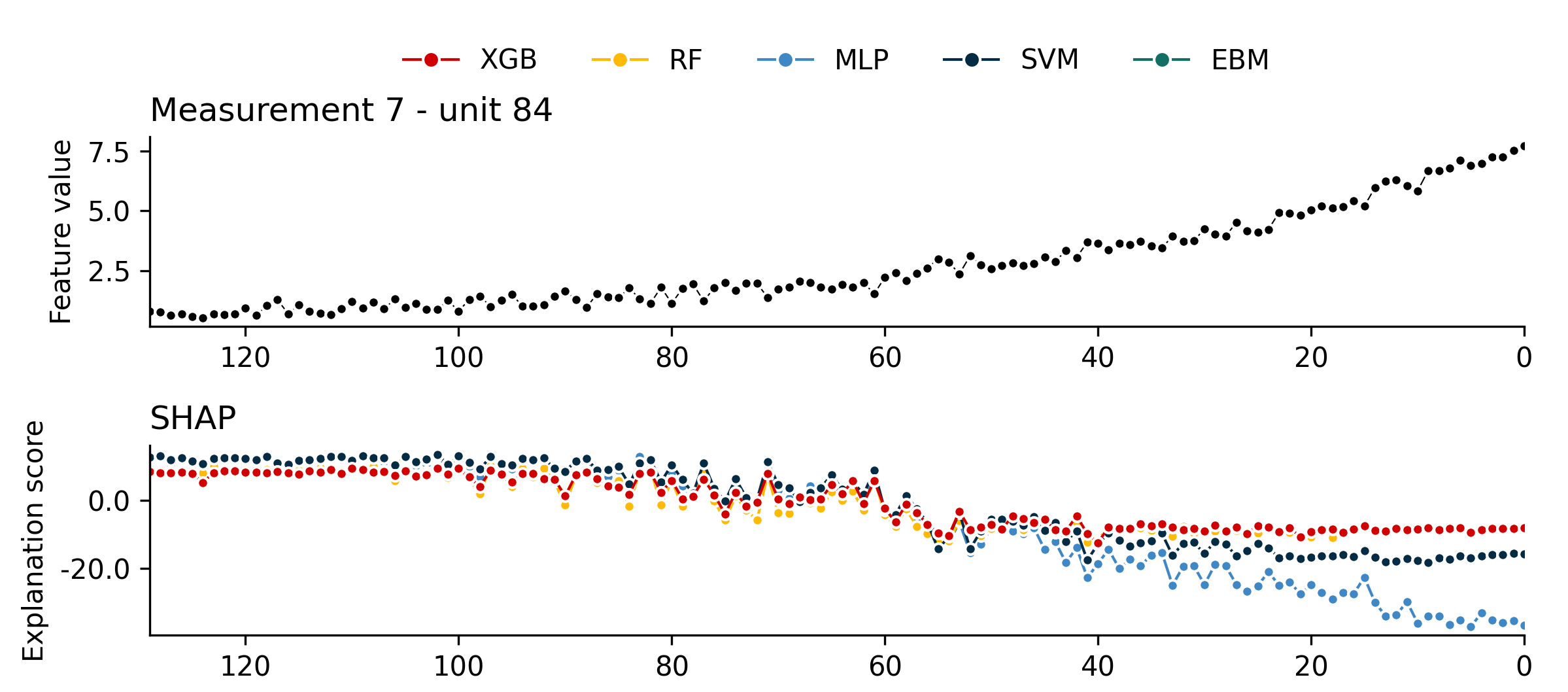}
    \caption{SHAP explanation scores for the selected feature using different ML methods. Adapted from~\cite{Jakubowski_2022}.}
    \label{fig:cmapss_shap_rul}
\end{figure}

Figure \ref{fig:cmapss_shap_rul}~\cite{Jakubowski_2022} presents the example of an issue which may be discovered by debugging the model with XAI. 
The presented measurement increases with the degradation of the component, but the explanation score for three out of four models saturate at a certain point, which is not logical.

Including XAI algorithms in the development pipeline raises the question of to what degree the explanations provided by the selected algorithm are reliable.
The XAI methods may fail to give the correct explanation, just like the model fails to provide an accurate prediction.
The evaluation of XAI algorithms themselves is another area of research~\cite{molnar2020interpretable}, but the developer of the PdM model should be aware of the problem.

With all their capabilities, XAI methods are a valuable extension of the PdM development pipeline, as they can show factors affecting the model predictions, which are hidden from the developer in standard model development steps.

\subsection{Decision Support}
Ultimately, predictive maintenance is just a single step towards a more efficient operation. Analysis and predictions do not make any impact unless better decisions and actions are made based on the obtained information. It is, therefore, crucial to evaluate explanations across their multiple dimensions, including understandability, trustworthiness, and usefulness in decision-making.

In particular, there are three key aspects to consider for improved PdM actionability: trust, severity analysis and complex repairs. Under all these three circumstances, the explanations play a vital role in allowing human decision-makers to consider the broader context. The different types of explanations can be used in the context-aware decision support \cite{Bobek2017UncertainCD} to close the loop in the decision-making process between humans developing and adjusting maintenance plans and black-box PdM techniques. 

First, in most industrial settings, it is crucial to establish a high level of trust and confidence for human experts to make complex decisions and long-term plans based on output from AI systems~\cite{fan2020transfer}. One of the roles of explainable machine learning methods is to understand predictions and their effect on the planning and optimisation tasks that human experts use to base their decision on. The expected efficiency benefits cannot be achieved if the repair technicians do not trust the PdM system and second-guess its predictions.

Second, not all problems are equally important. Beyond predicting the RUL for a particular component, a sound PdM system can provide additional context information~\cite{Calikus2022wisdom} concerning the severity or criticality of the current situation. A skilled technician can, from the explanations accompanying the machine learning models, get insight into additional consequences of the fault. For example, abnormal operation of one subsystem can cause collateral damage and further strain on other parts of the machine. With such extended understanding, an expert is able to make decisions improving overall efficiency without being locked into silo-based thinking.

Third, in some situations, the repair procedure is relatively simple and well-defined; a whole system or a single self-contained component is deemed faulty and should be replaced. In many cases, however, the relevant procedure is much more intricate. The complete repair plan and maintenance actions that must be performed based on the detected symptoms of damage and wear often require complex reasoning and planning processes; it involves many actors and balances different priorities~\cite{Galozy2022inf}.
Without explanations, the decisions made by complex, black-box machine learning models are often difficult for human experts to understand -- and, therefore, to act upon.


Today's primarily black-box AI does not provide these insights, nor does it support experts in making maintenance decisions based on the deviations it detects. What is more troublesome, though, is that the majority of XAI state-of-the-art solutions are not adequately addressing them either. The most popular approaches in XAI, by a huge margin, are those providing feature importance -- but they do not address \textit{any} of the needs listed in this section.
Ultimately, the effectiveness of the PdM system depends much less on the accuracy of the alarms the AI raises than on the relevancy of the actions operators perform based on these alarms. There is a need for a new generation of XAI approaches that recognise this fact.


\subsection{Product understanding and improvement} 
The degradation of the asset, in most cases, leads to a decrease in the machine's performance.
Parameters that quantify performance loss vary significantly depending on the use case.
In manufacturing, performance can be expressed in the form of, e.g., productivity (number of products in a certain period), quality (fraction of products which meet predefined requirements) and cost. 
In the automotive industry, key indicators would be uptime and energy efficiency.
High-quality maintenance is, without a doubt, one of the critical aspects of preserving or improving the machine's performance.
From such a perspective, explainable predictive maintenance models can help not only in scheduling the right maintenance actions, but also in improving the design of machines and processes. The potential long-term benefits are essentially endless.

Black-box models can be used to predict future equipment failures, the quality of products in a variety of manufacturing processes~\cite{Tercan2022ReviewQuality}, as well as in complex multistage processes~\cite{GenerativeNNQualityPrediction}.
The prediction itself may be helpful to improve product monitoring, but it does not allow one to understand and improve the process and design in the long term.
The real added value can be achieved by using XAI methods: the addition of the XAI layer or using interpretable models improves understanding of the black box model, thus understanding of the process and product, which can help in the overall improvement of performance~\cite{Senoner2022_XAI_Quality, Szelazek2021HRM}.
To a skilled technician or engineer, the XAI can potentially directly answer what should be improved in the process or point out the bottlenecks. 
Similarly, they can understand the mismatch between product design assumptions and actual usage in the field, leading to critical improvements and refinements in subsequent versions.


\section{Types of Explanations}
\label{sec:types}

The ultimate goal for XAI is to develop techniques that allow for models that are globally arbitrarily complex yet support sufficiently simple explanations for any desired purpose. This can be done either by specifying constraints that lead to models that have such properties inherently or as a \textit{post hoc} explainability layer. In this section, we discuss different techniques used to achieve this goal. Virtually all existing successful solutions, in practice, boil down to how the decision from the AI model that in itself is entirely incomprehensible for a domain specialist can be narrowed down into evidence that is specific to the scenario at hand. The latter is expected to be simple to understand, as it only contains a handful of the most-relevant conditions. As with all constrained models, of course, the benefits are inherently tied to the question of how well their design corresponds to the inherent properties of the data and problem at hand. Therefore, no single approach among those described below is inherently superior to the others; they are all applicable and valuable in different contexts.


To understand the AI explainability technical methods landscape \cite{arrieta2019explainable} proposes a taxonomy to distinguish models that are explainable by design \textbf{"interpretable-ml"} from opaque ones \textbf{"explainable-ml"} that require explanations through post hoc methods, either specific or model agnostic. These are further categorised by the explanation techniques, according to the authors, corresponding to the ``most common ways humans explain systems and processes by themselves.'' In the latter survey, authors suggest an alternative taxonomy, specific for Deep Learning methods \cite{8631448}, based on three concepts: how data is processed by the Deep Neural Network, how data is represented internally and how the model was specifically designed to simplify interpretation.

\cite{belle2020principles} attempt to bridge the available academic literature on AI explainability methods to educate industry data practitioners by creating a real-life scenario narrative (Loan Application). They ask different stakeholder questions concerning different methods, highlighting that there is no silver bullet: the best explanations result from combining different techniques.
\cite{Linardatos2021ExplainableAA} also organises XAI methods, using as a point of departure a mind-map taxonomy that additionally, recognizes fairness and sensitivity (robustness) as sub-domains of interpretability methods.
Alternatively, explainability taxonomy can be organized as an Engineering process that cuts through the different stages of development and deployment of an AI/ML system, as proposed by \cite{10.3389/fnins.2019.01346}. Potentially the combination of different development tasks can contribute to the overall capacity of the system's explainability; stage and goals correspond to 
pre-modelling, in-modelling, and post-modelling.

\subsection{Pre-modeling}
\label{sec:premodelling}
Pre-modelling explanations focus on understanding data through exploratory data analysis methods to draw insights, together with a domain expert, that can lead to the creation of new explainable features, describing data using standardisation documentation to support the interpretation of the domain and more effective usage of data as proposed by~\cite{gebru2021datasheets}, summarising data by leveraging the vast body of research on unsupervised methods either from early methods such as k-medoid clustering~\cite{article} or more recent methods such as ``Prototypes and Criticisms''~\cite{10.5555/3157096.3157352} that aim to find a minimal subset of representative samples. Data summarisation tasks (or other methods from the representation learning field) performed with a domain-specific expert can also drive the creation of new explainable features that can boost model explainability or explanations extracted by post hoc methods.

\subsubsection{Dataset summarisation}
Regardless of the higher-level goal (classification, clustering, regression, anomaly detection, etc.), the first modelling steps are always data preprocessing, including interesting patterns detection, data cleansing, missing values imputation, sampling, labelling, etc. 
These steps are not only about the statistical properties of data but DM/ML pipelines with multiple subgoals that contribute to the overall understanding of the data.
This process involves communication between domain experts, data scientists, technicians, and other stakeholders. 
Data scientists need to formulate and encode these tasks properly, 
which makes explainability an important aspect. 
Especially in industrial applications such as PdM, data are very often delivered unlabelled and poorly understood, and a proper analysis of the domain is crucial for designing high-quality systems.

Such understandability is achieved by EDA approaches, i.e., visualising the data, discovering patterns and anomalies, and confronting discovered knowledge with existing expert understanding.
In~\cite{pmlr-v162-crabbe22a}, the feature importance for unlabelled data is proposed, using XAI methods at the initial phase of data prepossessing to distinguish features that may play a key role in further ML tasks.
In~\cite{vispdm2021}, authors note that visual analytics-based approaches help maintenance log analysts extract and explain important patterns specific to certain issues. In survey~\cite{survey-vis}, the authors reported that 19\% of visual analytics in PdM have an explainability goal.
Along with data analysis and visualisation, another critical aspect of data summarising and understanding is conformance checking between existing domain knowledge and knowledge discovered from data. 
In~\cite{bobek2022knac}, authors proposed a guided process of verification and unification of knowledge obtained from experts and from data with clustering algorithms, to improve the quality of obtained clusters and their descriptions.
Similar concepts underlay~\cite{STEENWINCKEL202130}, where domain knowledge is utilised with data-driven anomaly detection to derive interpretable causes of machinery malfunctions.
Other, more generic methods for cluster analysis with XAI are also investigated. This includes tabular data description methods such as ~\cite{bobek2022clamp} as well as other data modalities such as time series~\cite{predmax2022}.

\subsection{In-modelling}
The goal of in-modelling explanations is to develop models that are intrinsically explainable, by selecting adequate algorithm complexity (and resulting performance) for the problem at hand, as argued by \cite{semenova2021study}.
%
The majority of modern machine learning algorithms are black boxes and cannot be interpreted inherently.
However, some techniques are considered interpretable, which means that by looking at the parameters and the structure of the model, one can determine all factors that lead to a certain decision.
The common examples of interpretable models are decision trees and rules, or linear regression. Decision trees are easily interpreted, as long as they remain relatively small since their design follows the natural process of human decision-making. 
Unfortunately, with simplicity (which induces interpretability), often comes relatively low predictive power meaning that these algorithms typically do not perform as well as state-of-the-art black boxes.

Linear regression, for instance, assigns each feature a weight, with the final prediction being the cumulative weighted sum. This way, one can easily determine the impact of each feature on the model output.
However, since linear regression is not able to model any non-linear relationships, it often fails on more complex tasks.
%
%
The key feature of these baseline linear models is that they are additive in nature, and, therefore, easy to explain how each variable is contributing to the final prediction. Predictive power can be increased by introducing models that have the capacity to learn non-linear relationships but keeping the additive properties, such as ``Generalised Additive Models,'' and finally by limiting interactions (two-way) with adapted GAM methods such as Explainable Boosting Machines \cite{https://doi.org/10.48550/arxiv.1909.09223}. Besides these approaches, options such as mechanistic models derived from scientific theories through mathematical equations are mentioned by \cite{miller2021breimans}. Learning cause-and-effect relations aims to explain the ``world'' instead of the ``mode,'' as proposed by \cite{10.5555/3238230}. Finally, an interesting and simpler approach towards causality is to condition a machine learning model with a known (causal) relationship implementing a monotonicity constraint, as suggested in \cite{10.1007/978-3-030-74251-5_2}.

\subsubsection{Physics-inspired models}
Physics-based modelling is an approach based on mathematical equations which describe a system's behaviour.
Such models may use no or little observed data to model the object of interest properly.
Physical models have a long history and were a topic of scientific interest for hundreds of years.
They can be expressed in the form of relatively simple relationships (e.g. heat balance in the oil tank) or very complex systems, often using differential equations (e.g. CFD~\cite{cfd} simulation of the temperature distribution within the oil tank).
In physics-based modelling, the level of detail must be properly selected for a given task.
Simple models are usually easier to understand and fast to compute, but they often over-generalise the problem.
On the other hand, complex physical models give very detailed and accurate answers but are usually difficult to understand and need a lot of computational resources, making them infeasible in real-world applications.

Predictive maintenance tasks usually apply to complex physical systems, where many aspects of the theoretical behaviour of the system are known beforehand, at the stage of system design.
Therefore, most components of interest can be partially described using physical formulations. At the same time, the complexity of the system makes it impossible to model its overall behaviour precisely.
As an example of a physical system, we consider a rolling mill, which reduces the thickness of the steel strip.
The physical equations, which govern the rolling process are well-established, and many of them were proposed as far as 100 years ago, e.g., Bland-Ford model~\cite{bland1948} for roll force and torque prediction.
This model makes several simplifying assumptions to the physics of the process, computing the roll force and torque within milliseconds.
Another approach to solving the same problem is to use the finite element method to numerically solve the differential equations describing the steel deformation process~\cite{rolling_fem}.
However, in practice, both models may fail to precisely predict the system behaviour due to uncertainties, which are present within the process.
In cold rolling, such uncertainties are mainly related to the tribology of the process and include factors like friction, lubrication and heat transfer~\cite{primer_tribiology}.
Each of the listed phenomena introduces a very high level of complexity to the system; therefore, it is infeasible to develop a PdM system purely based on physical equations.

Therefore, a hybrid approach which comprises both physics- and data-driven models is a promising compromise.
There are several ways in which data-driven models may be merged with physics-based models.
The physics model may be included as a regularisation parameter into the loss function of machine learning model~\cite{lake}, so that model is learning in parallel from data and physics.
In case of a low number of observations, which is often an issue in PdM problems, a transfer learning approach may be used -- data generated from the physical model is used to pre-train the deep learning model, and the fine-tuning is performed with real data~\cite{pinn_shroder}.
A data-driven model can be used to predict the residual error of a physics-based model, accounting for the imperfections of this model with ML~\cite{willard2022}.
In some cases, process variables that are not directly measured may be computed using available data and known physical relationships, as virtual sensors, to be fused with raw features and extend the input vector of ML models~\cite{pinn_fusion_phm}.
The ML model may also help in solving differential equations directly, decreasing the computational time needed~\cite{willard2022}.

In terms of explainability, hybrid models fall somewhere between black-box and glass-box models.
On the one hand, the physics incorporated into the model makes the predictions reliable and consistent with expert knowledge. On the other hand, the ML part of the model may still make it challenging to interpret the results.
In the case of complex physical models, which include many differential equations and variables, the inherent interpretability of the model is also not always necessarily very high. 
This is due to the fact that to understand the predictions of the system, human needs first to understand underlying physics and mathematical equations, which may not apply to non-qualified staff like maintenance technicians or managers.
However, the incorporation of physics into the PdM model is expected to increase the interpretability and reliability of the model, because the physics-based equations constrain the model to obey the prescribed physical laws.

\subsubsection{Causal models}
From the very beginning, the study of nature drove human knowledge, and knowing causality is often considered synonymous with understanding. In particular, the search for causes of natural phenomena animated several philosophers in Ancient Greece, such as Plato and Aristotle.
For instance, in~\cite{Plato}, the so-called ``inquiry into Nature'' consisted of a quest for ``the causes of each thing; why each thing comes into existence, why it goes out of existence, why it exists.'' Similarly, nowadays, the study of causality covers different disciplines, aiming to answer ``why?'' questions.

ML models are used in decision-making processes in real-world problems by learning a function that maps the observed features with the decision outcomes. However, these models usually do not convey causal information about the association in observational data, thus not being easily understandable for the average user, making it impossible to retrace the models' steps or rely on their reasoning. On the other hand, in their natural state, causal methods are highly explainable, as they mimic human reason. Moreover, this type of algorithm aims to explain a ``what if'' situation (if I change this, what will happen to that) instead of ``how'' (how would seeing this change my belief in that).

Causal models are designed to make explicit the {\it role and importance of each component of a machine learning model on its decisions with concepts from the causality} \cite{Moraffah2020CausalIF}, making them particularly suitable to answer specific questions about how certain aspects of the system influence others.
This concept is strictly related to the definition of causability (``the extent to which an explanation of a statement to a human expert achieves a specified level of causal understanding with effectiveness, efficiency and satisfaction in a specified context of use''~\cite{Holzinger19}). Since it represents the capability of a model to understand causal relationships and make them transparent to the user, 
it is a particularly relevant approach for PdM systems. These properties become especially important within a broader context of decision support systems; the ability of causal models to reliably describe the consequences of various alternatives makes them exceptionally enticing.

\subsection{Post-modelling}
The success of statistical ML models, such as DNNs, led many researchers to abandon interpretable-ml models~\cite{Rudin2019}. The proliferation of black-box solutions created incentives for explosive growth in research post hoc methods. Compared to the previous two stages, there is a proliferation of tools, often open source, that implemented post-modelling methods; in fact, the term ``XAI'' is sometimes used specifically in this context. The goal of the post-modelling stage is to extract explanations from previously developed models using different ideas, such as explaining a complex model by training a proxy ml-interpretable model \cite{molnar2019}, understanding the feature's relevance as in SHAP~\cite{SHAP} or LIME~\cite{LIME}, or describing model inner logic through data representative samples such as Counterfactual Explanations \cite{https://doi.org/10.48550/arxiv.2108.00783} or Influential Instances \cite{https://doi.org/10.48550/arxiv.1703.04730} methods.

\subsubsection{Explanation by example}
The example-based explanation is a general term for any method which provides a data instance to explain the prediction, like prototypical, adversarial, influential, or counterfactual examples. 
Prototypes are examples that best describe the data and work as representatives of the different classes in the distribution. On the other hand, criticisms are examples that any of the prototypes could not describe.

The k-medoids clustering method~\cite{Kaufman_clustering_1987} is the classical technique to find prototypes that are the most representative of the clusters in the data. Another method that finds both prototypes and criticisms is MMD-critic~\cite{Kim_examples_2016}; it finds prototypes as the examples that minimise the maximum mean discrepancy (MMD) between the prototypes distribution and the data distribution.
%
%
Moreover, the method selects the criticism examples as the instances that maximise the witness function, capturing data points at which the distributions of prototypes and data diverge.

In predictive maintenance, such prototypical examples can be representatives of different types of faults, and they can be used to draw insights about the signature of such faults in terms of feature values~\cite{kharal_explainable_2020}. Furthermore, prototypes can be the natural candidates to present the results of local explanation methods.
Prototypical examples can be useful as they provide an intuitive explanation of model decisions based on the similarities between the explained example and its closest prototype, which is similar to the human way of thinking. It can also help to provide an overview of the dataset based on a few most representative examples, which might help to identify some patterns in the data. On the other hand, because of their limited number, prototypes may not be able to convey a comprehensive picture of the variations in the dataset.

\subsubsection{Feature Importance}
Feature importance is by far the most popular method in the XAI research area~\cite{BHATT2019}.
The idea behind feature importance is to assign each input feature of the model a weight, which describes its impact on the model output.
In general, a value of feature importance close to zero means that there is a minor impact of the given feature on the model prediction, while high values indicate a significant impact on the prediction.
In some of the methods, positive and negative feature importance values might be assigned.
In such cases, the sign indicates in which direction the feature ``moves'' the prediction, e.g. whether the given feature leads to increasing or decreasing the RUL of the component.

Local Interpretable Model-Agnostic Explanations (LIME)~\cite{LIME} is one of the methods that has gained a lot of popularity.
The idea is to perturb the input instance to generate new observations in the neighbourhood of that point.
This new dataset is then used to train an interpretable model (e.g., linear regression) to assign feature importance values.
The key reasons for the popularity of LIME are that it is a model-agnostic approach, producing explanations for any black-box model, and it does not require access to training data to produce explanations.
One of the main drawbacks is the lack of stability; small changes in the data may result in large changes in the feature importance values~\cite{ALVAREZ2018}.
Thus, one can obtain vastly different explanations for very similar observations, which may be confusing.
Another popular method is Shapley Additive Explanations (SHAP)~\cite{SHAP}, inspired by the game theory work of Lloyd Shapley.
The general idea is to calculate the contribution of each feature to the prediction by building coalitions of features and randomly perturbing the data to see their impact.

\subsubsection{Sensitivity analysis}
Sensitivity analysis is used to determine how input features influence model outputs. This approach is similar to feature importance methods, with the difference in calculating the \textit{gradient} of the prediction for each input feature. Most methods assume input is an image (sometimes a time series), and they create a saliency map to provide insight into how neural networks make a prediction based on the input pixels (time steps).

Deconvolutional neural networks (DeconvNets) are a class of neural networks that operate in reverse. They are used to visualise the activations of individual neurons in the network and to identify the features the network has learned to recognise. By visualising the activations of individual neurons, it is possible to identify which features are most important in making predictions and to improve the interpretability of the network. As the max-pooling is not an invertible operation, this method produces Max Locations Switches to approximate the inverse of this operation.
Saliency is one of the first methods designed to visualise the input attribution of the Convolutional Network. Saliency is often associated with the whole method of displaying input attribution known as the saliency map, so this method is also called Vanilla Gradient or Grad. Based on the gradient of class c ($Y_{c}$) with respect to the input image I, this method generates the class saliency map.
Since grad has an issue with the flow of negative gradients, which decreases the accuracy of the higher layers we are trying to visualise,  Guided Backpropagation (GBP) is designed. Authors add a “guide” to the Saliency with the help of deconvolution to tackle this issue.

Class Activation Mapping (CAM) is another approach to producing a saliency map by replacing fully connected layers with convolutional layers and global average pooling (GAP). Due to this replacement, CAM requires to be fine-tuned and may reduce the performance, and it is not applicable for all types of networks. Also, it is applied on the last convolutional layer and does not care about previous ones. To solve these issues, Gradient-weighted Class Activation Mapping (Grad-CAM) uses the gradient to produce CAMs. There is no need for any modifications to the network to use this method, so the performance of the network does not change. This method is not limited to the last convolutional layer. Any of the convolutional layers in the network can be selected and generate the saliency map of that by computing the gradient.

Sensitivity analysis is primarily used in image processing; however,~\cite{GRADCAM_TS} proposed to use this approach for time series predictions, making CAMs and other derivatives a potential area of interest for the predictive maintenance community.

\subsubsection{Counterfactual Explanations}
One of the XAI methods that are natural to understand for humans is counterfactual explanations.
In its simplest form, a counterfactual explanation is a point in the feature space, which has a different label from the instance that we are trying to explain.
There are various restrictions on how to generate such a point, but the general idea is to find an example with the smallest possible distance from the original.
The purpose of such explanations is to answer the question, ``What is the minimum required modification to change the prediction of the model?.''
Since many PdM tasks are formulated as binary classification (healthy versus damaged asset) or regression (health indicator or RUL), counterfactuals are a promising solution.
In 2017,~\cite{wachter} first introduced counterfactual explanations as the method to understand the predictions of black-box models.
They formulated this as an optimisation problem in which we want to minimise the distance between original point $x$ and corresponding counterfactual $x'$ while ensuring the new prediction is 
any class different from the class of $x$.
%

One of the issues with counterfactuals is that many diverse explanations exist for the same observation, and the choice among them is highly dependent on the chosen distance metric.
In PdM, this could prove problematic, if -- for example -- we expect the counterfactual explanation to address the question of ``which component should be replaced to bring the system back to normal operation?''
An answer that could point to many different components, as some change in each of them could produce the desired output, is not particularly useful.
One solution has been introduced by~\cite{mothilal2020dice}, who proposed a method to generate a number of diverse explanations. 

The other issue connected with the generation of counterfactual examples is their feasibility, i.e., whether the proposed point is within the boundaries of the process.
An explanation that is an impossible state of the system will not be useful.
For example, if we monitor a car and notice an increased fuel consumption, we would like to know what are the potential reasons.
If the model learned that fuel consumption is low when the engine is turned off, the generated counterfactual could be that you keep the current speed but turn off the engine to lower the fuel consumption -- which certainly is not a desired result. Therefore, the feasibility of the explanation is one of the main concerns in the literature.
\cite{dandl2020} proposed to include a training dataset in the optimisation and to minimise the distance between the explanation and existing points.
A more versatile approach has been proposed by~\cite{mahajan2020preserving}, who included a Variational Autoencoder to ensure the feasibility of the generated explanations.


\section{Use Cases} \label{sec:usecases}
Finally, after presenting challenges and state-of-the-art solutions in the previous sections, we conclude the paper by providing concrete examples of PdM solutions in various domains. We share our experiences of applying Explainable Predictive Maintenance, which combines XAI and PdM, in four specific industrial use cases: commercial vehicles, metro trains, steel plants, and wind farms. We focus both on the practical applicability of existing solutions and on identifying open questions and areas requiring further research.
\subsection{Commercial Vehicles}

\begin{figure}[th]
    \centering
    \includegraphics[width=\textwidth]{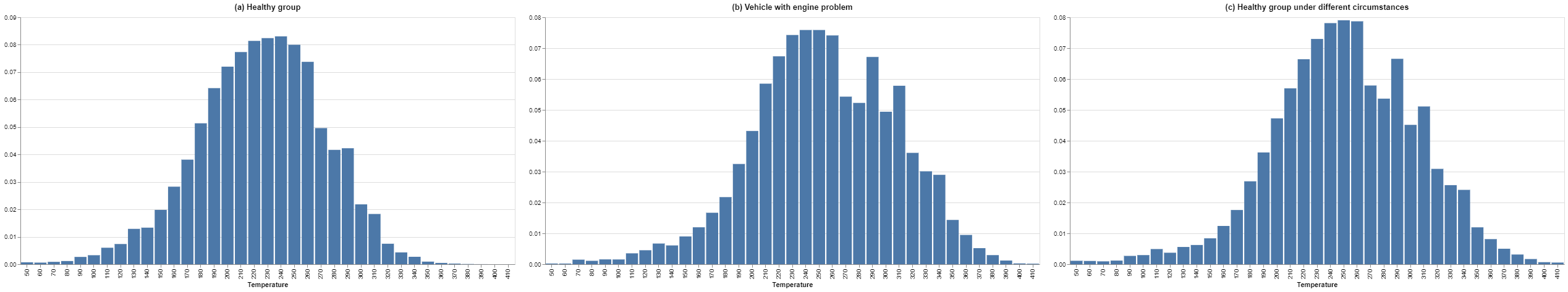}
    \caption{Distribution of engine temperatures of heavy-duty vehicles. The left and right panels show a healthy fleet under different operating conditions. The middle panel shows a vehicle with a cylinder problem -- clearly different from (a) but essentially impossible to differentiate from (c). The Bhattacharyya distance between the left and middle panels is $0.058$, but between the middle and right, it is only $0.002$.}
    \label{fig:volvoCylinder}
\end{figure}

Commercial heavy-duty vehicles, e.g., buses and trucks, are challenging domains for predictive maintenance. In contrast to production lines in factories with stable and well-controlled working conditions, vehicles operate in dynamic and harsh operating conditions~\cite{BOUGUELIA201833}. These conditions would deteriorate the system in different ways and may cause different types of faults, making the fault detection and prognosis process more difficult. One example demonstrating the difficulty is the distribution of engine temperatures, shown in Figure~\ref{fig:volvoCylinder}, where a fault (misaligned cylinder, middle panel) is clearly distinct from normal operation under one set of conditions (left panel) but almost perfectly hidden under other circumstances (right panel).

\begin{figure}[th]
    \centering
    \includegraphics[width=\textwidth]{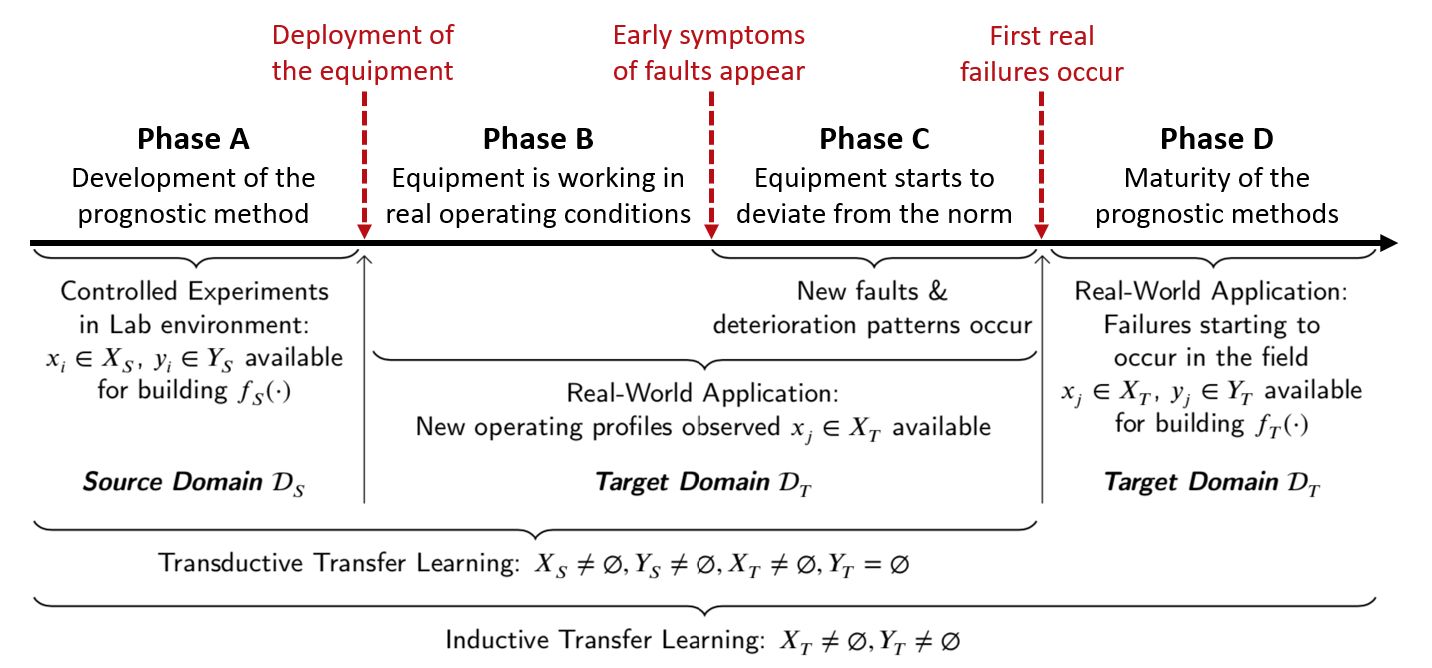}
    \caption{Transfer Learning setup for incremental development of PdM solutions in an industrial setting.}
    \label{fig:volvoTL}
\end{figure}

Moreover, modern vehicles are very complex mechanical systems with hundreds of thousands of ECUs onboard; narrowing down a problem and finding the root cause can be resource-consuming, e.g., numerous tests must be conducted, especially for novel ones.
The operation and success of commercial vehicles depend on the reliability of the system; due to the high cost of downtime, maximising the uptime is crucial.
Most diagnostic functions are developed using controlled experiments in the laboratories condition. However, equipment being deployed to the field might deteriorate in various ways, and unseen faults may occur under different operating conditions~\cite{Fan2016}. Therefore, a comprehensive strategy for continuous improvement of PdM solutions is required; see, for example, Figure~\ref{fig:volvoTL}.

Ensuring the safety of the operation, being able to proactively maintain the equipment (e.g., avoiding breakdown during the operation), and optimising planned downtime (e.g., reducing the time required for the preparation) requires more than detection of the faults and estimating the remaining useful life.
In the domain of commercial vehicles, XAI techniques can help utilise the output of various PdM tasks in real applications.

One can consider a system that monitors vehicles via streaming sensor data in a commercial fleet and notices whenever any vehicle is operating out of the ordinary~\cite{Bouguelia2018}. The explanation derived from the analysis (e.g. anomalous indicator in the signals that trigger the alarm and the component it corresponds to) can support the technician in the workshop for diagnosis, e.g., locating the faults in the systems, narrowing down the scope from various probable causes~\cite{Fan2015}. 
For example, if the engine temperature is outside the permitted range of the readings, engine parts or the cooling system can be prioritised for troubleshooting. 
Furthermore, domain experts can relate the diagnosis result to the explanation acquired, providing feedback on whether the system is analysing the suitable features and identifying new features that can be included for future data collection. This can potentially be based on post-mortem analysis of false positives or false negatives, e.g., what indicators were ``missed'' by the AI system and whether they are available in the current dataset. 

The objective for RUL prediction is to help the fleet manager schedule maintenance since it provides a timeframe on how long the system (or a component in the system) can operate according to the specification~\cite{app10010069}. The explanation, e.g., a list of factors or features that significantly impact the RUL, is helpful for the operator or technician with additional information on the system. The final maintenance decision should be made based on combined knowledge from AI and the human expert. Domain experts need to know what factors AI has already taken into account. For example, RUL prediction may be too optimistic if some factors in specific scenarios were overlooked, and an explanation to the expert can provide these relevant factors for further consideration.
The shape of the RUL prediction curves, i.e. trajectory patterns, might indicate different degradation processes of the equipment and might be closely related to the usage patterns. Any vehicle can fall into a particular group w.r.t. the wear pattern, and those groups can be described in different ways. The explanation can be made by identifying important factors for the degradation speed.

\subsection{Metro Trains}
\begin{figure}[tb]
    \centering
    \includegraphics[width=1.0\textwidth]{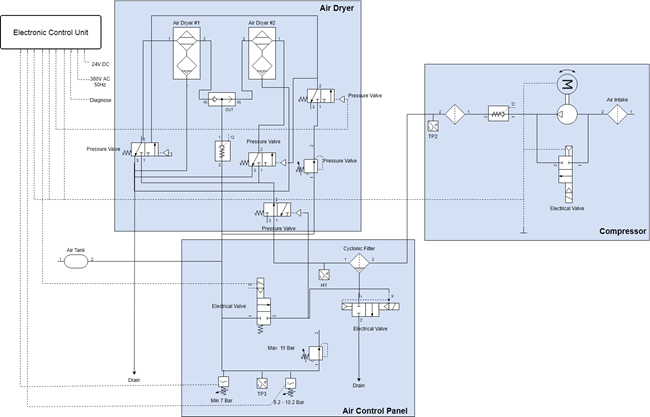}
    \caption{The Air Production Unit system with the position of the main sensors.}
    \label{fig:Metro}
\end{figure}

The occurrence of failures in public transport vehicles during their regular operation is a source of numerous problems, especially when they cause interruptions in the trip. The negative impacts affect not only the operator company but especially the clients, who are disappointed with their expectations of trust in transportation. In this context, the early detection of such failures can avoid the cancellation of trips and the withdrawal of service from the respective vehicle and thus is of enormous value. 

The compressed air production Unit (APU) in the Metro do Porto (MP) EuroTram trains is one of the pieces of equipment that most contribute to the cancellation of trips. Its primary function is the production of compressed air for supply to four ``clients,'' namely: 1) coupling system; 2) lubrication; 3) sandpits; 4) secondary suspension. 
The latter is responsible for keeping the height of the vehicle level, regardless of the number of passengers on board. This is a highly requested element throughout the day,  and the absence of redundancy causes its failure to result in the immediate removal of the vehicle for repair.
This case study aims to detect failures in the compressed air production system early to prevent them from occurring during commercial operation, thus reducing the loss of trips caused by breakdowns and consequently increasing the reliability of rolling stock.

Using RCM (Reliability Centered Maintenance), FMEA (Failure Mode and Effects Analysis), and FMECA (Failure Mode, Effects, and Criticality Analysis) analysis, it was possible to identify the APU weaknesses. 
Using the fault history, it was then possible to identify the most common faults and their associated causes. From this base, the most critical variables to be monitored were identified, estimating the values between which they fall during operation and the installation locations where these quantities can be measured were defined.

A system of acquisition, concentration, and transmission of data was defined so that the information read in real-time in the sensors and transducers could be centralised in a server. The sensors are placed in three subsystems: compressor, control panel, and several segments of the compressed air pipeline (including air-drying towers and a cyclonic filter). The signal acquisition system installed in the two APUs of two vehicles (APU01 and APU02) collects data from eight sensors (pressure, temperature, and electric current consumed, placed in different components of the APU) and eight digital signals collected directly from the APU control unit. 
Figure~\ref{fig:Metro} presents the schema of the APU, indicating the main components and the position of the sensors.

The data acquisition rate is 1~Hz, and the information is sent to the remote server every 5 minutes using the GSM network. The data collection of the two units began on 12 March 2020 and is continuously operational to date. Every day, for each APU, a report is generated with the information on the sensor signals.

\begin{figure}[bt]
\centering
\includegraphics[width=.6\linewidth]{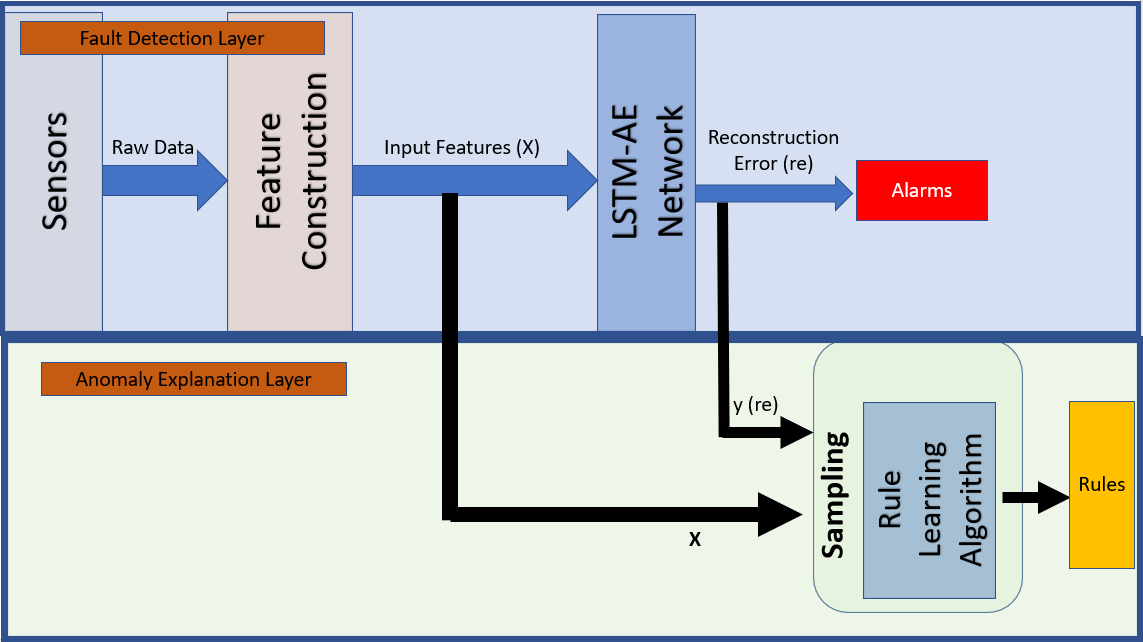}
\caption{The Neural-Symbolic Online Anomaly Explanation System. The top panel details the fault detection system, while the bottom panel details the explanation system. Both systems run online and in parallel.}
\label{fig:fig0}
\end{figure}

Figure~\ref{fig:fig0} presents the global architecture of the proposed neuro-symbolic explainer for rare events \cite{ribeiro2023online}.
It works online, receiving real-time data from sensors installed on the train. 
The figure details the two layers of the system: the {\em fault detection} and {\em explanation} layers.
The main component in the {\em detection layer} is the LSTM-AE network trained with examples from the normal behaviour of the machine we want to monitor. Each new observation is propagated through the autoencoder network. 
The reconstruction error is computed as the mean square difference between the input and the output:  
$re = 1/n \times \sum_{i=1}^n(x_i-\hat{y_i})^2$, where $n$ is the number of features, $x_i$ the input features, and $\hat{y_i}$ the corresponding output. When the reconstruction error, $re$ exceeds a threshold, we signal an alarm, meaning that the input is not from the system's normal behaviour.  
The second layer, the {\em explanation layer}, receives as independent variables ({\bf $X$}) the input features of the LSTM-AE network, and the dependent variable ($y = re$) is the corresponding reconstruction error. It learns regression rules using the AMRules algorithm~\cite{DuarteGB16}, mapping $y=f(X)$. Both layers run online and in parallel, which means that for each observation $X$, the model propagates the observation through the network and recomputes the reconstruction error ($y=re$). The second layer updates its model using $\langle X,y \rangle$. If $y > threshold$, the system signals an alarm with an explanation of the rules that trigger, for example, $X$.
The first layer is the failure detection layer and is unsupervised. The second layer is supervised and explains the failure in terms of the input features.
High extreme values of the reconstruction error ($re$) are a potential indicator of failures.
This architecture allows two explanations: i) Global level: the set of rules learned that explains the conditions to observe high predicted values, and ii) Local level: which rules triggered for a particular input.

We present below a subset of rules obtained for the Train dataset when using 
ChebyOS+AMRules.
Rules 0 and 1 are related to an oil leak on the motor of the compressor. 
Larger values of sensor {\tt dig7} less oil in the system. The output of the rules indicates the severity of the failure. 
\begin{verbatim}
Rule 0: If dig7 > 2258.0000 Then 219.1732
Rule 1: If dig7 > 2187.0000 Then 42.8249
\end{verbatim}

Rule numbers 2, 3 and 4 are related to an air leak located after the pneumatic control panel. Rule 5 is related to failure on the compressor control system, i.e., the escape valves installed on the APU are opened when the compressor is trying to fill the tanks.

\begin{verbatim}
Rule 2: If B1_TP3 > 7345.6000 and B5_MC > 1925.7000 Then 1.8116
Rule 3: If dig8 > 251.0000 Then 2.3932
Rule 4: If B6_TP3 < 5635.1000 Then 2.4445
Rule 5: If B2_H1 > 378.1000 Then 1.8791
\end{verbatim}

A sample of the data from this case study is publicly available and is described in~\cite{veloso2022metropt}.

\subsection{Steel Plant}
\begin{figure}[ht]
     \centering
     \begin{subfigure}[b]{0.8\textwidth}
         \centering
         \includegraphics[width=\textwidth]{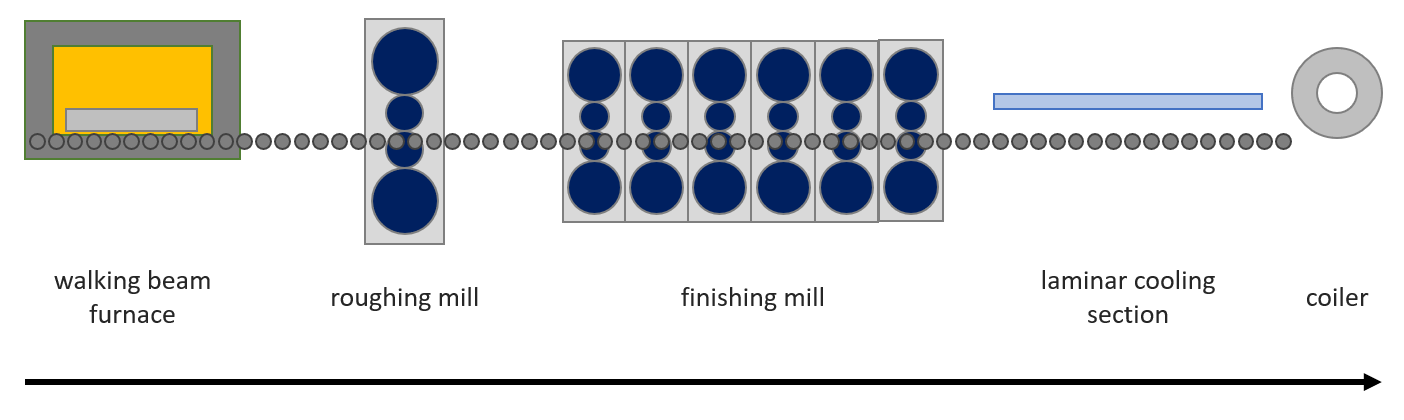}
         \caption{Hot rolling mill~\cite{Jakubowski_2021}}
         \label{fig:hrm_flow}
     \end{subfigure}
     \hfill
     \begin{subfigure}[b]{0.5\textwidth}
        \centering
        \includegraphics[width=\textwidth]{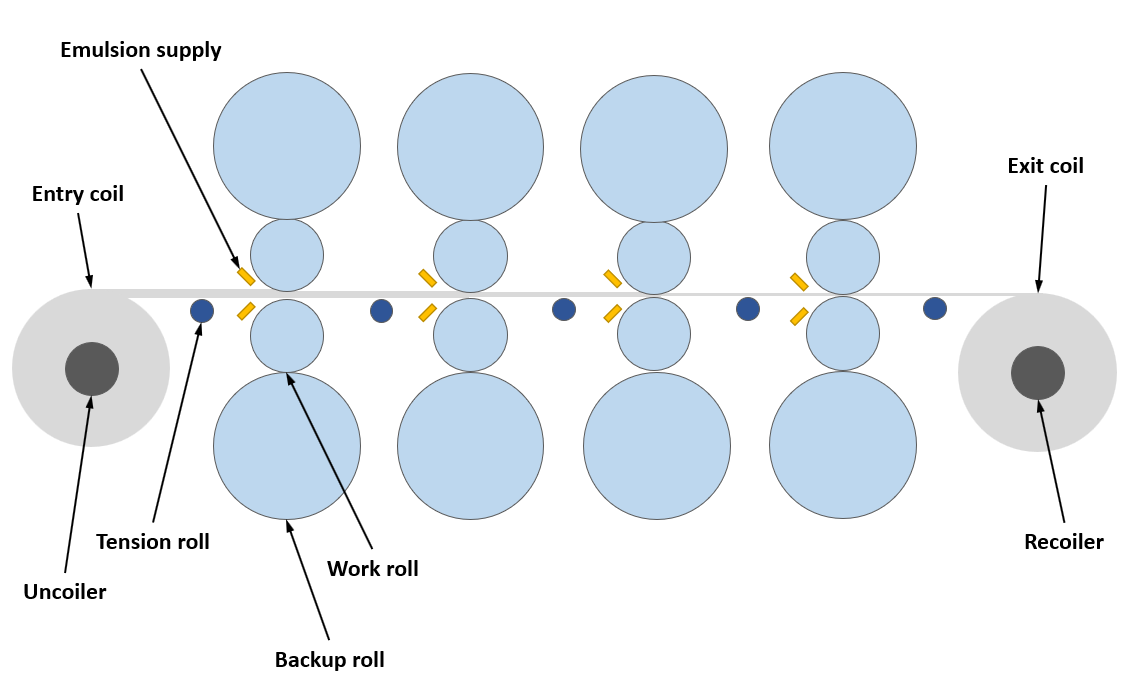}
        \caption{Cold rolling mill~\cite{Jakubowski_2022_dsaa}}
        \label{fig:crm_flow}
     \end{subfigure}
        \caption{Simplified process flow in rolling mills.}
        \label{fig:steel_flow}
\end{figure}

The steel factory use case includes the application of data-driven predictive maintenance approaches in strip rolling mills of ArcelorMittal Poland.
Rolling is one of the key steps in the steel manufacturing process, which determines the physical properties of steel, as well as its quality.
After the continuous casting process, steel is about 200mm thick; the aim of the flat rolling process is to reduce this dimension significantly -- even below 1mm.
In ArcelorMittal Poland, there are two plants dedicated to strip rolling: a hot rolling mill (HRM) and a cold rolling mill (CRM). 
Firstly, a steel slab passes through HRM, where the thickness is reduced to approximately 2.0 - 25.0 mm. 
If a lower thickness is needed, the steel strip is cold rolled.
Figure \ref{fig:steel_flow} presents a simplified scheme of each rolling process in the use case.

One of the crucial pieces of equipment installed in HRM and CRM is the work rolls, which are used to reduce the thickness of the strip.
They have a significant impact on the performance of the process and the quality of the steel produced.
They are in direct contact with the steel; therefore, any defect in these rolls can easily be transferred to the surface of the strip, leading to the production of downgraded material.
The roll wear process is relatively fast, and each pair is replaced more than once a day.
After replacement, the rolls are recovered in the workshop and put back into production, but the cost of recovery is significant.
Currently, rolls are replaced in a preventive way after a pre-defined mileage or time.
However, better control of the roll degradation process could result in longer production campaigns and a consequent increase in productivity.
A lower frequency of replacement of the work roll would also reduce maintenance costs.
In addition, intelligent wear monitoring could help reduce the number of defective products due to an early indication of the need for a roll replacement.
Through data-driven predictive maintenance, a health indicator could be determined for the work rolls, which could be used to support decisions for the replacement of the work rolls.

Another application of ML methods in the predictive maintenance of rolling mills is anomaly detection.
Rolling mills are complex systems, where the operation of one rolling stand may impact the behaviour of the other stands.
Therefore, monitoring the process is a difficult task, and advanced algorithms, which learn the normal behaviour of the mill, would be a valuable tool for decision support.
However, the black-box nature of most ML models would drastically reduce the usefulness of such a monitoring system.
The anomaly detection system in such a scenario should be trained not only to detect the anomalous measurements, but also to indicate the measurements which lead to this state.
For this second purpose, Explainable AI techniques are needed, as most of the state-of-the-art anomaly detection models are black boxes.

So far, we have developed an unsupervised model for the detection of anomalies in the hot rolling process using the LSTM Autoencoder~\cite{Jakubowski_2021} and a semi-supervised approach with the use of Variational Autoencoder~\cite{Jakubowski_2021_Sensors}, with the aim of distinguishing between low and high wear of work rolls.
We have also developed a model to predict the wear of work rolls in cold rolling, which utilises the physics of the process to increase the accuracy of the mode~\cite{Jakubowski_2022_dsaa}.
In terms of the model explainability, we have utilised the SHAP method and Counterfactual Explanations, to build an understanding of the decisions of the developed models.
The explanations were able to indicate which features caused the anomalous state of the mill or work rolls, helping to find the root cause of the anomaly.

\subsection{Wind Farms}
Like every other complex and heterogeneous system, wind turbines, a principal part of wind farms, are prone to faults that can affect their performance and increase maintenance costs. Moreover, the impact of different component faults on wind turbines is different. To mention just a few: some failures result in this system shutting down or even turbine destruction. Some other defaults are less severe and result in a malfunction of the turbine systems, decreasing power conversion performance.

\begin{figure}[ht]
\centering
     \includegraphics[trim=0cm 0cm 0cm 0cm, clip=true, width=14cm, height=10cm, scale=0.6]{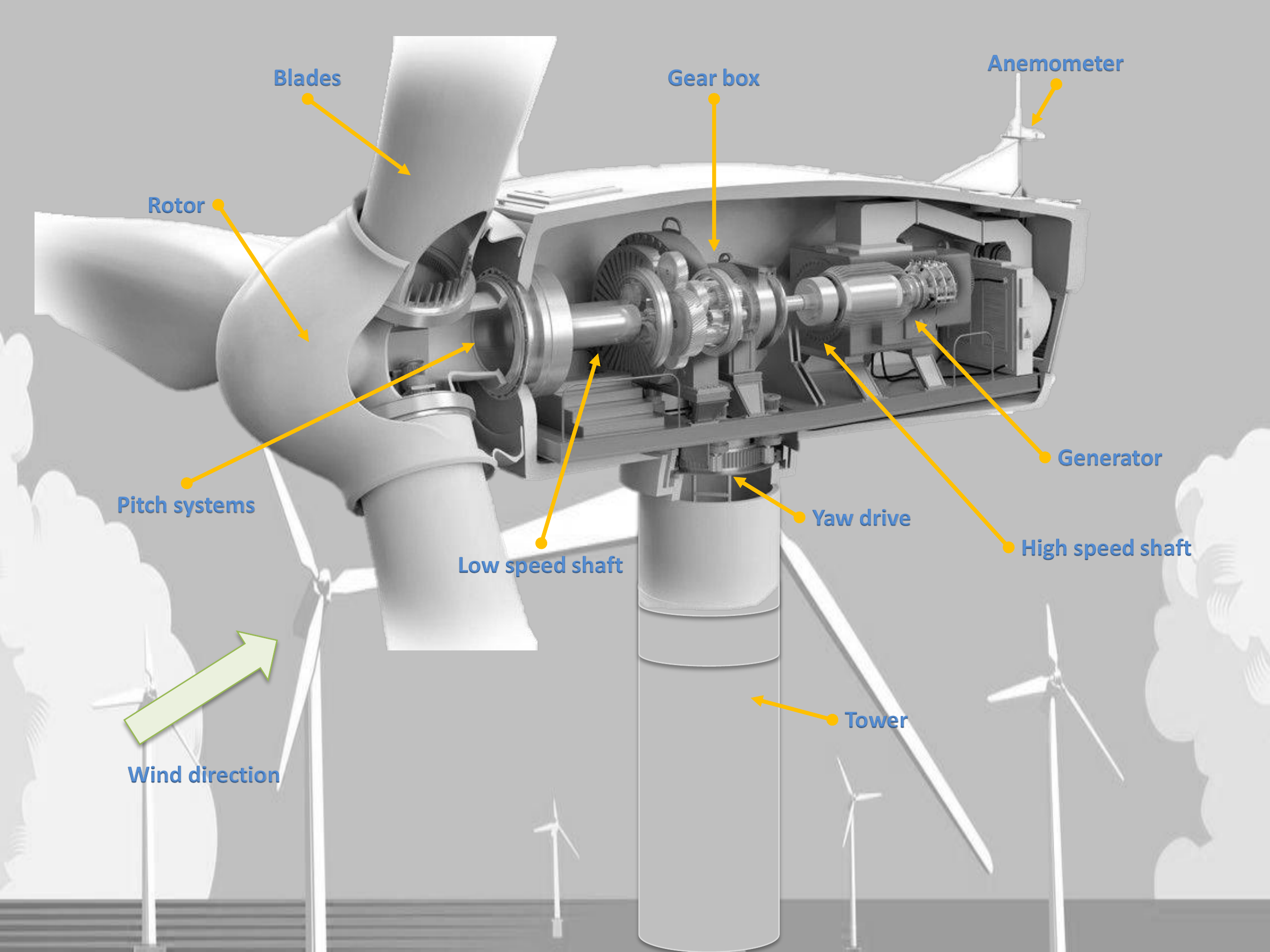}
    \caption{Wind turbine component scheme}
    \label{fig:dilem}
   \end{figure}
   
Let us give more examples. Pitch actuator faults can be difficult to identify and explain due to the large variability in wind speed, confusion between faults and noise, and outliers. In addition, the consequences of the failures can be hidden because the controller can compensate for the effects of faults. Faults in the hydraulic system (Figure \ref{fig:dilem}: Gearbox part) can be due to pressure drop or air content in the oil. The pressure drop is generally due to a blocked pump or leakage in a hose. Actuator faults in the power converter of the generator (Figure \ref{fig:dilem}: Generator part) system may be due to internal causes. Defects in semiconductor devices are major internal causes of the power converter fault. Failure of sensors used to measure blade pitch angle, blade rotating speed, wind speed, generator speed, generator torque, etc., may be due to electrical and mechanical causes (Figure \ref{fig:dilem}). These faults can result in decreased control performance since the feedback signals in the wind turbine control loop are provided by these sensors. Also, gear tooth damage can be due to offset and eccentricity of tooth wheels, and so on.
 
In other words, detecting and isolating faults in wind turbine systems are critical tasks. In addition, it should be done as fast as possible to restore optimal operating conditions in the shortest time. Meanwhile, human operators should quickly understand the actions to perform. That is, the design of a suitable decision-making system rises to support the operator in its maintenance task \cite{lrbib02}. It could help operators to understand failure severity, occurrence, and detection rating, and decide accordingly the best sequence of actions to do \cite{gradient}. In other terms, the decision support system should produce detailed information and description reports by gathering and analysing data. It should explain the causes of wind turbines and perform a maintenance plan concerning (i) the failure mode, (ii) the degradation severity, (iii) the dynamic evolution of failures and (iv) the spare parts inventory needed to replace the faulty components.

We present in Table \ref{tb:Rules} below a subset of rules obtained for the wind turbine dataset to explain the occurrence of anomalies in the system. We can see that some features, such as the active power generator, blade pitch angle and average water temperature of the voltage circuit protection of the sub-components of the generator, are more interesting to watch. Indeed, the implemented recommender and diagnoser systems identified them as the most important features, allowing us to explain the origin of the problem in the physical system. By analysing the obtained rules given the anomaly dynamic and the global evolution of the monitored system behaviour, it could be possible to help the maintenance manager in his maintenance choices and adapt the optimal repair and maintenance strategy among several alternatives created according to different priorities.

\begin{table}
\caption{Identified rules for anomaly detection for wind turbine systems. $K_{ur}$ and $M_f$ define, respectively, the Kurtosis and Margin factor values of each feature. Features number 8, 14 and 15, which are equivalent, respectively, to the blade pitch angle, the latest average production of the active power generator and the average water temperature of the voltage circuit protection. }
\label{tb:Rules}
\begin{center}
\resizebox{0.99\textwidth}{!}{
\begin{tabular}{|c||c|c|}
\hline
Fault type & Descriptions & Identified rules\\
\hline
\hline
{$F_1$} & High temperature for generator bearing 
 & $(K_{ur}(feat_8)>40)$ and $(K_{ur}(feat_{14})>40)$ and  $(K_{ur}(feat_{15})>40)$
\\
{$F_2$} & High temperature of Transformer
& $(K_{ur}(feat_{15})>40)$ and $(K_{ur}(feat_{15})<=100)$ and $(I_f(feat_{14})>25)$ and $(M_f(feat_{14})<=500))
$\\
{$F_3$} & Oil leak
 & $(M_f(feat_{14})>15)$ and $(M_f(feat_{14})<=25) $ and $(K_{ur}(feat_{15})>100)
$\\
{$F_4$} & Bearing generator damage &$(M_f(feat_{14})>25)$ and $(M_f(feat_{14})<=500)$ and $(K_{ur}(feat_{15})>100)
$\\
{$F_5$} & Generator damage
 & $M_f(feat_{14}) > 500$\\
\hline
\end{tabular}}
\end{center}
\end{table}

\section{Conclusions}
\label{sec:conclusions}


We believe this paper can be very engaging and relevant for two separate audiences: (i) PdM professionals who can learn more about XAI and how it can benefit them by replacing black-box models for benefits, and (ii) XAI professionals who can use PdM as an application area and testbed for their ideas. We hope that our work can bring those two communities closer together. Complex systems typically have highly interconnected designs, where sensor readings depend not only on the state of a monitored component but also on the state of all the surrounding components. Therefore, capturing the interactions between components and subsystems is crucial for the explanations to be meaningful. The use of explanations beyond the scope of the AI/ML model that is generating them, and their connection to broader context or downstream tasks, is a novel but also challenging perspective for XAI researchers.

Throughout this paper, we have established that XAI is very important for PdM. At the same time, there is a lot of potential for further research, since current methods are falling short of addressing all the practical challenges.
Key scientific contributions of this work are the taxonomy of PdM tasks and the corresponding XAI solutions; the matching of various types of explanations to different purposes and different PdM tasks. Another aspect is the difference between explaining the data (and thus the real-world generating this data) versus explaining the model, which is under-studied in XAI literature. Taking the well-known loan application as an example: a satisfactory counterfactual explanation is any that makes the model flip the decision from ``deny'' to ``approve.'' In some cases, it may -- somewhat counterintuitively -- involve taking additional loans; the evaluation is not based on the underlying reality (the ability to pay off the loan) but only on the model's outputs. The PdM setting is significantly more challenging: a counterfactual that increases RUL \textit{prediction} without increasing \textit{actual} RUL is useless at best, and probably actively harmful.

Most of the research on explainability in the field of predictive maintenance has been limited to calculating feature importance for the different tasks to understand the model's outcome. However, those explanations are rarely helpful in practice.
In addition, most explainability methods are designed to work in supervised settings. At the same time, there are very few works on unsupervised methods, such as anomaly detection, which is the most widely used approach in the area of predictive maintenance. 
%
Another critical question is whether to create a black-box, high-accuracy model first and add explainability in a post hoc fashion, or to build inherently interpretable models instead -- there is currently heated debate on this topic in the scientific literature, with a clear need to evaluate both approaches to fairly compare and contrast them over a set of representative PdM tasks.

Another challenge is how to use explanations to support different actors in designing the proper intervention for the monitored system. Since the ultimate aim is to make a plan of action that avoids imminent failure or restores normal operation, explanations are not only used to provide insights to human operators but also used as inputs to a decision support system. For human experts to make complex decisions and long-term plans based on output from AI systems, it is crucial to establish a high level of trust and confidence. This topic has been studied for a long time, but still, it is more art than science. A critical aspect of PdM is to have a robust methodology to evaluate explanations across their multiple dimensions (e.g. understandability, trustworthiness, and usefulness in decision-making). 
A methodology that can, in a flexible way, combine both quantitative engineering and performance measures, as well as qualitative domain-knowledge-driven criteria formulated by human experts, is far from consensus in the scientific community.



\section*{Acknowledgements}
The paper is funded from the XPM project funded by the National Science Centre, Poland under CHIST-ERA programme (NCN UMO-2020/02/Y/ ST6/00070), French National Research Agency (ANR) under CHIST-ERA programme (ANR-21-CHR4-0003), Swedish Research Council under grant CHIST-ERA-19-XAI-012 and Portuguese Funding Agency,
FCT - Fundação para a Ciência e a Tecnologia under CHIST-ERA programme (CHIST-ERA/0004/2019).
The research has been supported by a grant from the Priority Research Area (DigiWorld) under the Strategic Programme Excellence Initiative at Jagiellonian University. 

\bibliography{sample}

\end{document}